\documentclass{article}

\usepackage{microtype}
\usepackage{graphicx}
\usepackage{subcaption}
\usepackage{booktabs} 
\usepackage{multirow} 
\usepackage{enumitem}
\usepackage{hyperref}

\usepackage[preprint]{icml2026}

\usepackage{amsmath}
\usepackage{amssymb}
\usepackage{mathtools}
\usepackage{amsthm}
\usepackage{enumitem}
\usepackage{listings}
\usepackage{xcolor}

\definecolor{codegray}{RGB}{60,60,60}
\definecolor{codebg}{RGB}{248,248,248}

\lstdefinelanguage{json}{
  basicstyle=\ttfamily\footnotesize,
  showstringspaces=false,
  breaklines=true,
  frame=single,
  rulecolor=\color{codegray},
  backgroundcolor=\color{codebg},
  keywordstyle=\color{codegray},
  stringstyle=\color{codegray},
  commentstyle=\color{codegray},
  numbers=none
}

\usepackage[capitalize,noabbrev]{cleveref}

\theoremstyle{plain}

\theoremstyle{definition}

\theoremstyle{remark}

\usepackage[textsize=tiny]{todonotes}

\icmltitlerunning{ADEPT: RL-Aligned Agentic Decoding of Emotion via Evidence Probing Tools}

\begin{document}

\twocolumn[
\icmltitle{
  ADEPT: RL-Aligned Agentic Decoding of Emotion via Evidence Probing Tools
  \texorpdfstring{ ---\\}{ --- }
  From Consensus Learning to Ambiguity-Driven Emotion Reasoning
}



  \icmlsetsymbol{equal}{*}

  \begin{icmlauthorlist}
    \icmlauthor{Esther Sun}{cmu}
    \icmlauthor{Bo-Hao Su}{cmu}
    \icmlauthor{Abinay Reddy Naini}{cmu}
    \icmlauthor{Shinji Watanabe}{cmu}
    \icmlauthor{Carlos Busso}{cmu}
  \end{icmlauthorlist}

  \icmlaffiliation{cmu}{Language Technologies Institute, Carnegie Mellon University, Pittsburgh, PA, USA}

  \icmlcorrespondingauthor{Esther Sun}{esthers@andrew.cmu.edu}


  \icmlkeywords{Machine Learning, ICML}

  \vskip 0.3in
]



\printAffiliationsAndNotice{}  

\begin{abstract}
Speech Large Language Models (SLLMs) enable high-level emotion reasoning, but often produce ungrounded, text-biased judgments without verifiable acoustic evidence. In contrast, SSL encoders such as WavLM yield strong acoustic representations yet remain opaque discriminative models that offer limited interpretability. To bridge this gap, we introduce the Agentic Decoding of Emotion via Probing Tools \textbf{(ADEPT)}
framework, which reframes emotion recognition as a multi-turn inquiry process rather than a single-pass prediction. ADEPT transforms an SLLM into an agent that maintains an evolving candidate set and adaptively invokes dedicated semantic and acoustic probing tools within a structured pipeline of candidate generation, evidence collection, and adjudication. Crucially, ADEPT enables a paradigm shift from \emph{consensus learning} to \emph{ambiguity-driven emotion reasoning}. Since human affect exhibits complexity and co-occurrence of emotions, we leverage minority annotations as informative signals instead of discarding them as noise. Finally, we integrate Group Relative Policy Optimization (GRPO) with the Evidence Trust Gate to explicitly couple tool-usage behaviors with prediction quality and enforce evidence-based reasoning. Experiments demonstrate that ADEPT improves in most cases the primary emotion accuracy while substantially improving minor emotion characterization, producing explanations grounded in auditable evidence.
\end{abstract}

\section{Introduction}
The deployment of Speech Emotion Recognition (SER) systems in high-stakes settings, such as automated customer service, mental health screening and empathetic human--computer interaction, increasingly demands not only accurate predictions but also transparent and auditable reasoning \citep{Nfissi2024UnveilingHiddenFactors,Jordan2025SpeechMentalHealth, Li2025EMORL}.
 In practice, the value of SER often lies in explaining \emph{why} an emotion is perceived, especially when affective states are subtle, mixed, or socially masked \cite{Sethu2019AmbiguousEmotion, Liu2023SpeechEmotionPerception}. For example, an utterance labeled as ``Neutral'' may reflect genuine calmness, polite suppression of distress, or the coexistence of mild anxiety and composure \cite{Singh2023SpeechAttentionSER}. Such cases highlight a central property of human affect: \emph{complexity} and \emph{co-occurrence} \cite{Larsen2001HappySadSameTime}, where primary and minor emotions can coexist within the same spoken segment \cite{Mower_2009, Chou2022ExploitingCooccurrence,Chou2024AmbiguitySubjectivitySER}. However, most SER research remains anchored to plurality-based consensus supervision \cite{Chou2024AmbiguitySubjectivitySER}. Given multiple annotators, the prevailing pipelines collapse diverse perceptual judgments into a single winner-takes-all label, implicitly treating minority votes as annotation noise \cite{Sethu2019AmbiguousEmotion}. This simplification suppresses the inherent ambiguity essential for dynamic reasoning, thereby preventing models from leveraging the minor emotions
that are consistently present in human perception \cite{Devillers2005ChallengesRealLife, Su2025ReasoningBeyondMajority}. Even soft-label approaches typically focus on distributional smoothing, but often fail to explicitly represent the semantic dependencies among co-occurring emotions \cite{Fayek2016SubjectivenessSER,Lotfian_2017,Kang2025MultilabelConversationalERC}. Moreover, consensus-based filtering further exacerbates the problem by discarding high-disagreement samples, resulting in substantial data waste \cite{Lotfian2019CurriculumSER,wu-etal-2023-dont} and removing precisely the examples that expose emotional complexity \cite{Chou2025RevisitingModelingEvaluation}. We refer to this phenomenon as the \emph{Consensus Paradox}: the more a dataset enforces agreement, the less it preserves the perceptual variability that defines real-world affect \cite{kenyon-dean-etal-2018-sentiment, Niu2024RethinkingEmotionAnnotations}.

The consensus paradox challenge is compounded by a structural gap between signal modeling and reasoning. On the one hand, Self-Supervised Learning (SSL) \cite{Yang2021SUPERB, Nfissi2024UnveilingHiddenFactors} speech encoders such as WavLM \cite{Chen2021WavLM} capture strong acoustic representations and achieve high accuracy under conventional supervision, yet they largely operate as opaque discriminative models with limited interpretability \cite{Pasad2022ComparativeSpeechAnalysis, Jayasinghe2025InterpretabilityExplainabilitySER}. On the other hand, Multimodal Large Language Models (MLLMs) offer high-level natural language reasoning and can articulate plausible emotional narratives, but often struggle to ground their judgments in fine-grained, verifiable acoustic evidence \cite{Tang2023SALMONN, Bai2024HallucinationMLLM, Hu2024WavLLM}. As a result, current systems either achieve strong recognition performance without trustworthy explanations, or produce explanations without reliable evidence alignment \cite{ Pavlick2023SymbolsGroundingLLM, Jayasinghe2025InterpretabilityExplainabilitySER}. Importantly, grounding emotion reasoning requires more than attaching static summaries of either acoustic or semantic features. When ambiguity arises, the system must be able to actively query targeted evidence, including localized pitch excursions, energy bursts, and relevant linguistic cues, to verify competing hypotheses through measurable and specific evidence retrieval. \cite{LarrouyMaestri2025SoundEmotionalProsody, Han2025BenchmarkingBridgingEmotionConflicts}.

Speech Emotion Recognition (SER) should be reframed not merely as a representation learning problem, but as an emotion reasoning problem under ambiguity: a system must preserve competing interpretations and justify its decision with \emph{auditable, hypothesis-specific evidence} retrieved from the signal. This framing calls for \emph{sequential verification}, which actively tests hypotheses via targeted semantic and acoustic measurements, rather than one-shot classification or post-hoc narrative generation. To this end, we introduce the \textbf{A}gentic \textbf{D}ecoding of \textbf{E}motion via \textbf{P}robing \textbf{T}ools (\textbf{ADEPT}) framework, which transforms an MLLM into a multi-turn agent that maintains an evolving candidate set and adaptively invokes structured \emph{semantic} and \emph{acoustic} probes. The resulting tool forms an auditable evidence trail that supports principled adjudication of both primary and minor emotions.

ADEPT further enables a shift from \emph{consensus learning} to \emph{ambiguity-driven emotion reasoning}: instead of collapsing annotator disagreement into a single label, we leverage minority votes as informative supervision for emotion co-occurrence. Finally, we integrate \textbf{Group Relative Policy Optimization (GRPO)} \cite{Shao2024DeepSeekMath} with an evidence trust gating mechanism to couple prediction quality with evidence-seeking behavior, encouraging the agent to acquire informative and reliable evidence before committing to claims. Experiments show that ADEPT improves in most cases the primary emotion accuracy while substantially enhancing the recovery of co-occurring minor emotions and produces explainable, evidence-grounded reasoning trace.

Our contributions are:
\begin{itemize}
    \item We propose \textbf{ADEPT}, a multi-turn agentic SER framework that performs policy-driven evidence acquisition via structured semantic and acoustic probing tools.
    \item We integrate \textbf{GRPO} with an evidence trust gate to align tool-augmented reasoning with prediction quality, preventing non-informative tool use while yielding evidence-grounded rationales and strong performance across co-occurring emotions.
    \item We advocate \emph{ambiguity-driven emotion reasoning} by treating minority annotations as supervision for co-occurring primary and minor emotions, rather than noise to discard.
\end{itemize}

\section{Related Work}

\textbf{Paradigms in SER: From Signal to Semantics.}
Recent SER research largely follows two diverging paradigms, resulting in a persistent signal--semantics gap. 
The first paradigm is the \emph{acoustic signal-centric} modeling, dominated by self-supervised speech encoders such as HuBERT \cite{Hsu2021HuBERTMaskedPrediction}
 and WavLM \cite{Chen2021WavLM}. 
These models excel at extracting fine-grained paralinguistic cues from raw audio and achieve strong recognition accuracy, yet they are typically deployed as opaque discriminative predictors with limited capacity to explicitly reason about the selected emotional class. \cite{Pasad2022ComparativeSpeechAnalysis,ma-etal-2024-emotion2vec}. 
In contrast, the second paradigm is the \emph{generative MLLM paradigm}, which leverages the cognitive and linguistic capabilities of large language models; MLLMs can move beyond classification toward richer affect understanding.
Progress in this direction has been rapid, including instruction-tuned emotion reasoning models such as Emotion-LLaMA \cite{Cheng2024EmotionLLaMA}
 and EmoLLM \cite{Yang2024EmoLLM}
, generative and empathetic systems such as AffectGPT \cite{Lian2025AffectGPT} and BLSP-Emo \cite{wang2024blspemo}, and adaptation and augmentation strategies that inject acoustic information into LLMs via parameter-efficient tuning (e.g., EmoSLLM \cite{Thimonier2025EmoSLLM}) or synthesized acoustic descriptions (e.g., AA-SLLM \cite{Mai2025AASLLMAA}).
However, despite improved semantic expressiveness, most existing approaches remain fundamentally \emph{passive}: acoustic evidence is provided as embeddings or static summaries rather than being actively queried and verified.
This limitation becomes critical under ambiguity, where models may produce plausible yet physically ungrounded explanations when the transcript and acoustic content conflict.

\vspace{0.5em}
\textbf{Subjectivity, Disagreement, and Minor Emotions.}
Emotion perception is inherently subjective and rarely admits a single gold-standard label \cite{Devillers2005ChallengesRealLife}, making annotator disagreement a defining property of real-world emotion datasets \cite{Lotfian2019CurriculumSER, Chou2024AmbiguitySubjectivitySER}.
Prior work shows that plurality-vote and majority-vote aggregation can induce substantial information loss by suppressing minority annotation \cite{Tavernor2025ModelingAnnotatorsSER}. Annotations that do not agree with the consensus class can still reflect valid perceptual variations conveyed in the speech \cite{Chou2025RevisitingModelingEvaluation}. We refer to the annotations that
deviate from the consensus class as minor emotions. Recent efforts have addressed this issue through soft-label learning \cite{Sridhar2021GenerativeSoftLabels}, uncertainty modeling \cite{Chou2022ExploitingCooccurrence, Kang2025MultilabelConversationalERC}, and ensemble-based mitigation \cite{Fayek2016SubjectivenessSER, Uma2021LearningFromDisagreement}, which explicitly accounts for annotator inconsistency.
While these methods statistically acknowledge ambiguity, they primarily aim to \emph{reduce} or \emph{smooth} disagreement \cite {Chou2024AmbiguitySubjectivitySER,Kang2025MultilabelConversationalERC}.
Consequently, the potential of annotator variation as informative supervision for disentangling and recovering co-occurring primary and minor emotions remains largely unexplored, as it is still commonly treated as noise.

\vspace{0.5em}
\textbf{Agentic Frameworks and Reasoning Optimization.}
Motivated by agentic AI, recent works introduce structured reasoning workflows for emotion understanding, including hierarchical or chain-of-thought reasoning, as exemplified by Agent-MER \cite{Lai2025AgentMER} and Affective-CoT \cite{Huang2025AffectiveCoT}.
In parallel, reinforcement learning has been explored to align Audio-Language Models with emotion-related objectives and improve reasoning faithfulness, such as EMO-RL \cite{Li2025EMORL, Lian2025AffectGPTR1}.
However, most existing methods still focus on high-level semantic planning or label optimization, and do not formulate emotion understanding as a multi-turn, interactive evidence acquisition process. When ambiguity arises, the system should be able to jointly leverage semantic consistency verification and low-level acoustic probing to contrast competing emotion candidates. The agent should be able to autonomously decide whether to invoke semantic or acoustic tools based on uncertainty and the source of conflict, thereby achieving verifiable evidence alignment.

\section{Label Construction}
\label{subsec:label_construction}

\begin{figure}[t]
    \centering
    \includegraphics[width=0.95\linewidth]{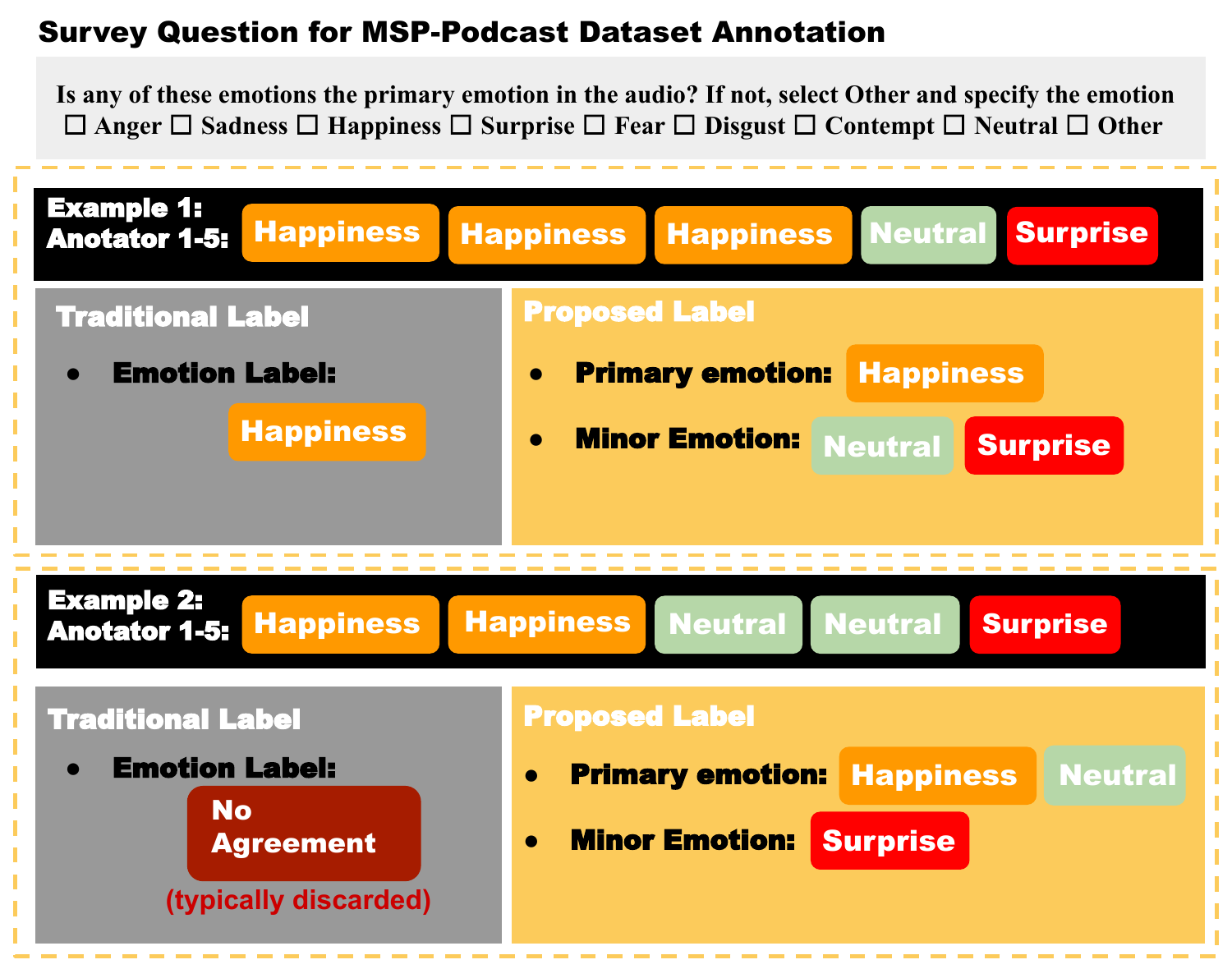}
    \caption{Illustration of our flexible label construction strategy for the annotations in the MSP-Podcast corpus, preserving both tied primary emotions and minority-vote minor emotions.}
    \label{fig:label_construction}
\end{figure}

In this study, we utilize the categorical emotion annotations from the \textbf{MSP-Podcast corpus V2.0 }\cite{Busso2025MSPPodcast}, where each audio segment is evaluated by a minimum of five annotators. The annotation protocol requires raters to answer a forced-choice question: ``Is any of these emotions the primary emotion in the audio? If not, select Other and specify the emotion''\footnote{See Appendix \ref{sec:appendix_msp} for more details about the corpus.}. The candidate set includes eight standard emotional categories---Anger, Sadness, Happiness, Surprise, Fear, Disgust, Contempt, and Neutral---plus an \emph{Other} option. To derive ground truth while mitigating data loss from traditional ``tie-breaking'' discards, we employ a flexible Plurality-based Consensus strategy (Figure~\ref{fig:label_construction}). Unlike strict majority rules, our approach explicitly models ambiguity: (a) \textbf{Primary Emotion} is defined as the class(es) receiving the highest vote count; crucially, in tie cases (e.g., equal votes for \textit{Happiness} and \textit{Neutral}), we retain all tied classes as valid primary labels to preserve mixed dominant states instead of filtering them out; (b) \textbf{Minor Emotion} includes any standard category selected by the annotators that is not included in the primary emotion(s), we preserve the classes to capture subtle affective nuances; and (c) for a closed-set taxonomy, we restrict labels to the eight standard categories, excluding responses marked as \emph{Other}.

\begin{figure*}[tb]
    \centering
    \includegraphics[width=0.8\textwidth]{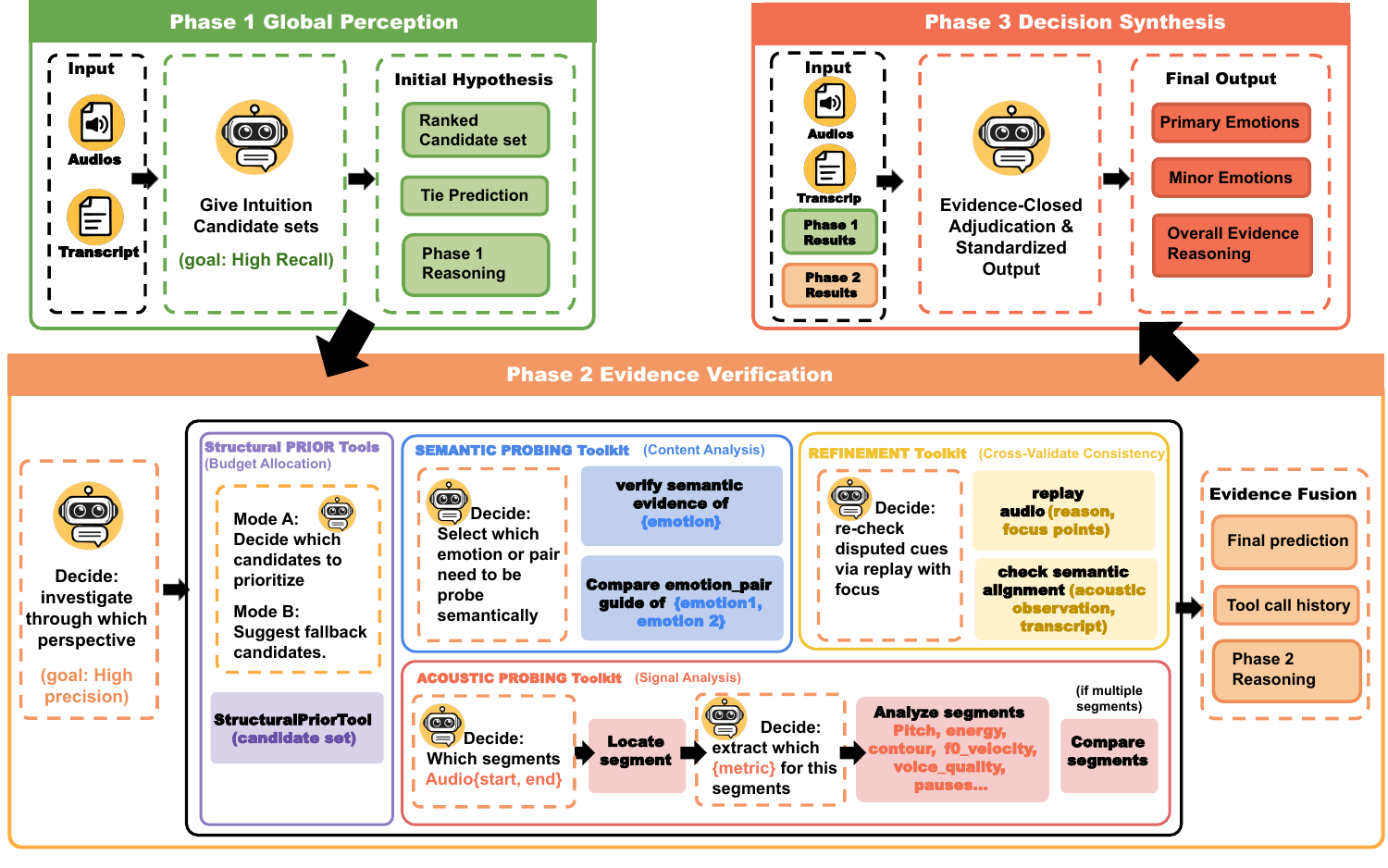}
    \caption{\textbf{ADEPT three-phase inference workflow with tool-mediated evidence probing.}
\textbf{Phase 1 (Global Perception)} performs high-recall hypothesis initialization from audio and transcript, producing a ranked candidate set, tie prediction, and coarse reasoning.
\textbf{Phase 2 (Evidence Verification)} executes adaptive evidence accumulation under a soft budget, combining four complementary tool families:
(i) \emph{STRUCTURAL PRIOR} for budget allocation and fallback candidate scheduling;
(ii) \emph{SEMANTIC PROBING} for span-level verification and pairwise disambiguation of confusable emotions;
(iii) \emph{ACOUSTIC PROBING} for emotion-neutral, localized signal measurements via a coarse-to-fine routine (\emph{locate} $\rightarrow$ \emph{analyze} $\rightarrow$ \emph{compare}); and
(iv) \emph{REFINEMENT} for closed-loop re-checking (e.g., replay with updated focus points) when evidence is conflicting or insufficient.
\textbf{Phase 3 (Decision Synthesis)} performs \emph{evidence-closed adjudication}, prohibiting further tool calls and synthesizing only Phase-2 retrieved observations to produce standardized outputs (primary emotions, minor emotions, and overall evidence-grounded reasoning).}
\vspace{-6pt}
    \label{fig:adept_three_phase_workflow}
\end{figure*}

\section{Proposed Framework: ADEPT}
\label{sec:adept}

We propose the \textbf{A}gentic \textbf{D}ecoding of \textbf{E}motion via \textbf{P}robing \textbf{T}ools (\textbf{ADEPT}) framework, an agentic solution that operationalizes \textbf{Explicit Information Retrieval (EIR)} as a \emph{structural constraint} for emotion reasoning under annotator disagreement, grounding predictions in verifiable evidence rather than free-form generation. Figure~\ref{fig:adept_three_phase_workflow} illustrates the proposed three-phase ADEPT framework.

Unlike one-shot SER classification, ADEPT transforms an MLLM into a multi-turn reasoning agent that maintains an evolving candidate set and adaptively invokes semantic and acoustic probing tools to resolve ambiguity (e.g., primary--minor co-occurrence or top-vote ties). The inference process follows a structured three-phase pipeline comprising candidate generation, evidence collection, and adjudication. To prevent hallucination and ensure auditability, the system strictly separates evidence from interpretation: tools convert raw signal measurements into standardized observations, which serve as the sole basis for the agent's reasoning. In the following sections, we detail the core components of this framework. First, we outline the execution flow of the \textbf{Three-Phase Inference Pipeline} (Section~\ref{sec:three_phase_pipeline}), followed by a formal definition of the \textbf{Evidence Probing Toolkit} (Section~\ref{sec:toolkit_design}) that powers these reasoning steps.

\subsection{Inference Protocol: The 3-Phase Pipeline}
\label{sec:three_phase_pipeline}

As shown in Figure \ref{fig:adept_three_phase_workflow}, ADEPT formalizes SER as a hierarchical reasoning process.
By temporally separating hypothesis formation, active evidence acquisition, and final adjudication, the framework mitigates confirmation bias while retaining the flexibility of agentic exploration.

\paragraph{Phase 1: High-Recall Hypothesis Initialization.}
The inference begins with a \emph{High-Recall Initialization}.
Given multimodal input $(x,t)$, the agent synthesizes coarse audio--text cues to construct a broad initial candidate set $\mathcal{C}_0$.
Crucially, Phase~1 is designed to \emph{preserve ambiguity}: it performs ranking and prioritization rather than hard filtering, yielding a sufficiently large top-$K$ candidate set that tolerates uncertainty and supports downstream evidence-driven refinement.

\paragraph{\textbf{Phase 2: Adaptive Evidence Accumulation.}}
This phase operates as a \textit{conditional computation loop} driven by the evolving ambiguity of the candidate set $\mathcal{C}_t$.
Departing from rigid execution graphs, the agent dynamically orchestrates the \textbf{Evidence Probing Toolkit} (Sec.~\ref{sec:toolkit_design}) to maximize information gain under a soft budget constraint.
The execution logic is governed by three integrated mechanisms:

\textbf{a. Prior-Driven Scheduling \& Recovery.}
During Phase~2, the agent leverages the \textit{StructuralPriorTool} to optimize evidence acquisition. Instead of brute-force exploration, it uses pair-level scheduling signals to prioritize the most informative hypothesis comparisons under a limited budget. Crucially, if accumulated evidence invalidates the current pool, this mechanism enables \emph{controlled backtracking} by suggesting minimal candidate expansions, recovering from Phase-1 bottlenecks without combinatorial explosion.

\textbf{b. Theory-Grounded Evidence Extraction.}
To acquire verifiable observations, the agent actively selects between \textit{Semantic} and \textit{Acoustic Tools}.
This process is a structured interrogation rather than a feature lookup:
semantic probing follows a \textit{Schema-Guided Protocol} (identifying appraisal factors such as valence polarity or agency),
while acoustic probing employs a \textit{Dynamic Perception Strategy} (using alignment-based anchoring and dual-reference bucketing).
This approach ensures that all extracted features are isomorphic to the agent's linguistic focus and interpretable.

\textbf{c. Closed-Loop Refinement via Bidirectional Replay.}
Under evidence conflict or uncertainty, the agent performs bidirectional re-checking by replaying audio or re-validating pragmatic text cues, without introducing new evidence types.
This approach prevents premature commitment when semantic and acoustic signals disagree.
The refinement step updates the verification state by re-auditing the disputed cues, yielding more consistent evidence for Phase~3 adjudication.

\begin{figure}[t]
    \centering
    \includegraphics[width=\linewidth]{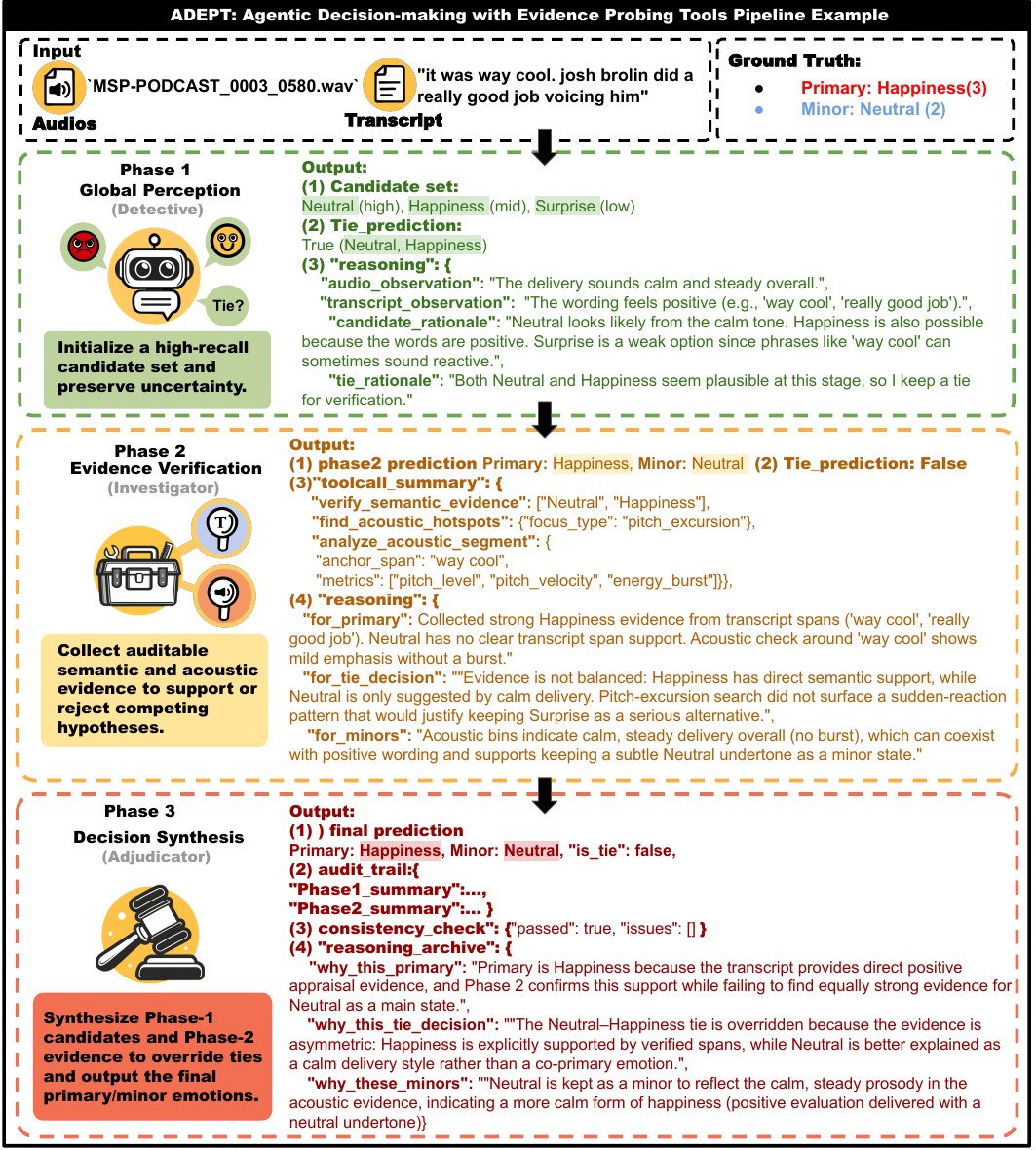}
    \caption{\textbf{ADEPT three-phase inference protocol (illustrated example).}
    Phase~1 initializes a high-recall candidate set and preserves uncertainty (e.g., a predicted tie).
    Phase~2 collects auditable semantic spans and localized acoustic evidence via tool calls to support or reject hypotheses.
    Phase~3 performs evidence-closed adjudication, overriding ties and outputting the final primary/minor emotions with an explicit audit trail.}
    \label{fig:adept_three_phase_example}
\end{figure}

\paragraph{Phase 3: Evidence-Closed Adjudication.}
In the terminal phase, the action space is restricted to reasoning operations only, prohibiting further information retrieval.
Crucially, to prevent confirmation bias, the \textit{Structural Prior} is explicitly excluded from this phase.
The agent functions as a judge, synthesizing the evidence accumulated in Phase~2 (semantic spans and acoustic bins) into calibrated predictions.
This \emph{evidence-closed} constraint ensures decision integrity, guaranteeing that all final judgments are causally traceable to explicit, verifiable tool observations rather than hallucinated latent features or training-set priors. Figure~\ref{fig:adept_three_phase_example} provides a concrete walkthrough of this end-to-end reasoning process.

\subsection{Toolkit Design: Action Space for Evidence Probing}
\label{sec:toolkit_design}

To enable evidence-driven reasoning, we define a structured action space $\mathcal{A}$ comprising four tool families. Unlike black-box classifiers, these tools function as extraction operators returning verifiable observations (e.g., transcript spans, discretized signal bins) rather than latent labels.

\textit{(i) Structural Prior Tools (Efficiency via Scheduling).}
StructuralPriorTool is a lightweight scheduling operator that leverages corpus-level co-occurrence statistics to optimize the navigation path through the hypothesis space. Derived from training-set statistics (see Appendix~\ref{sec:appendix_prior}), this tool is used exclusively to support efficient evidence acquisition in Phase~2, rather than to predict emotions.
Concretely, it provides two types of scheduling signals:
(i) \textit{Budget Allocation}, which prioritizes verification among highly confusable candidate pairs; and
(ii) \textit{Adaptive Backtracking}, which suggests statistically linked alternatives when the current hypothesis set is rejected.
By confining the prior to search navigation, ADEPT improves exploration efficiency while preventing any leakage of decision evidence into the final phase.

\textit{(ii) Semantic Probing Tools (Schema-Guided Verification).}
SemanticProbingTool is a theory-grounded verification operator that validates emotion hypotheses against the transcript to extract verifiable textual evidence. The tool implements a schema-guided protocol grounded in Scherer’s \textit{Component Process Model} (CPM) \cite{Scherer2001Appraisal}, instantiated using the \textbf{7-factor appraisal schema} validated by \citet{Tak2025AwareYetBiased}. Rather than open-ended generation, it adopts a literal-first strategy that maps appraisal factors into concrete linguistic cues and requires extraction of verbatim text spans, rendering semantic verification falsifiable and reducing hallucination on short, context-free utterances (see Appendix~\ref{sec:appendix_semantic}).

\textit{(iii). Acoustic Probing Tools (Auditable Signal Measurements).} AcousticProbingTool is a signal extraction operator that accepts raw audio and transcript-aligned boundaries to output discrete, linguistically interpretable signal observations. To quantify \emph{how} it is said, the agent employs a \textit{Dynamic Perception Strategy} to localize salient regions and retrieve evidence.
Crucially, we implement a Dual-Reference Bucketing mechanism to bridge the Numeric--Semantic Gap. Rather than raw scalars, the tool returns interpretable descriptors (e.g., \texttt{High Pitch}, \texttt{Volatile Energy}) relative to both global (speaker-level) and local (utterance-level) reference distributions.
The agent requests a minimal subset of psychoacoustic correlates (e.g., prosody, voice quality, spectral balance) to maximize information gain (See the full feature list in Appendix~\ref{sec:appendix_acoustic_tools}).

\textit{(iv). Refinement Tools (Cross-Modal Adjudication).} RefinementTools are context-aware adjudication operators that re-evaluate existing semantic and acoustic evidence to resolve cross-modal inconsistencies during Phase~2. This family functions as a \textit{safety valve} for closed-loop correction when accumulated evidence is inconsistent or inconclusive (e.g., sarcasm, where positive semantics contradict negative prosody). Rather than extracting new feature types, Refinement tools support \emph{bidirectional re-injection} to resolve dissonance: (i) Acoustic-conditioned Text Re-check, which re-evaluates the \textit{pragmatic meaning} of a transcript (e.g., detecting irony) given the observed emotional tone; and (ii) Semantic-conditioned Audio Replay, which directs the agent to re-attend to specific signal segments guided by textual cues. This tool prevents premature commitment by integrating conflicting signals into a coherent judgment.


\subsection{Optimization via Group Relative Policy Optimization (GRPO)}
\label{sec:grpo_optimization}

We optimize ADEPT with Group Relative Policy Optimization (GRPO)~\citep{Shao2024DeepSeekMath} to improve protocol-compliant tool use under the Explicit Information Retrieval (EIR) constraint.
Given an input $(x,t,r)$, we sample a group of $K$ trajectories $\{\tau_k\}_{k=1}^{K}$ from $\pi_{\theta_{\text{old}}}$ and update $\pi_\theta$ using clipped policy optimization with Kullback--Leibler (KL) divergence regularization.
Each trajectory spans the full ADEPT protocol (Phase~1 candidate initialization, Phase~2 evidence probing, and Phase~3 evidence-closed adjudication), where tool outputs are treated strictly as observations.

\noindent\textbf{Reward Components.} To align correctness with structured reasoning behavior, we define a composite reward as a weighted sum of five verifiable components, with weights empirically selected based on pilot experiments.
Specifically, the components enforce \textbf{structural constraints} via output executability ($R_{\text{fmt}}$) and pipeline integrity ($R_{\text{phase}}$), while optimizing \textbf{reasoning quality} through prediction accuracy ($R_{\text{out}}$), evidence grounding ($R_{\text{evid}}$), and strategic probing ($R_{\text{tool}}$). Detailed reward formulations are provided in Appendix~\ref{app:reward_details}. The resulting composite reward is formalized in Eq.~\ref{eq:composite_reward}, where $\tau_{\text{EIR}}(\tau)$ is an Evidence Trust Gate that is defined below.
\begin{equation}
\begin{aligned}
R(\tau)
&= 0.5 \, R_{\text{fmt}}
+ 0.2 \, R_{\text{phase}}
+ 0.4 \, R_{\text{out}} \\
&\quad + \tau_{\text{EIR}}(\tau)\Big(0.2 \, R_{\text{evid}} + 0.3 \, R_{\text{tool}}\Big).
\end{aligned}
\label{eq:composite_reward}
\end{equation}

\noindent\textbf{Evidence Trust Gate.}
To prevent the policy from exploiting evidence-related rewards without improving prediction correctness, we introduce an Evidence Trust Gate $\tau_{\text{EIR}}(\tau)\in(0,1]$ that modulates the contribution of evidence and tool rewards in Eq.~\ref{eq:composite_reward}.
The gate is designed to activate evidence-based supervision only when evidence-seeking behaviors are systematically more informative for correct trajectories than for incorrect ones. To compute this gate, we first define an \textit{auxiliary} evidence score $S_{\text{evid}}(\tau)$ that summarizes evidence-seeking behaviors within a trajectory:
\begin{equation}
S_{\text{evid}}(\tau)
=
\lambda_{\text{e}} R_{\text{evid}}(\tau)
+
\lambda_{\text{t}} R_{\text{tool}}(\tau),
\end{equation}
where $\lambda_{\text{e}}=0.2$ and $\lambda_{\text{t}}=0.3$ follow the weights used in Eq.~\ref{eq:composite_reward}.
This score is not optimized directly; instead, it serves as an internal statistic for assessing the reliability of evidence-related actions. Let $\mathcal{G}^{+}$ and $\mathcal{G}^{-}$ denote the sets of correct and incorrect trajectories within the same rollout group, respectively, and let $\mu^{+}$ and $\mu^{-}$ be the mean evidence scores over $\mathcal{G}^{+}$ and $\mathcal{G}^{-}$.
We then compute the Evidence Trust Gate as:
\begin{equation}
\tau_{\text{EIR}}(\tau)
=
\begin{cases}
1, & \text{if } \mu^{+} \ge \mu^{-},\\
\exp(\mu^{+}-\mu^{-}), & \text{otherwise}.
\end{cases}
\label{eq:evidence_trust_gate}
\end{equation}

This mechanism ensures that evidence-related rewards contribute fully only when correct trajectories exhibit stronger evidence-seeking behavior than incorrect ones, thereby discouraging non-informative or reward-hacking tool use.

\section{Experiment}
\subsection{Experimental Setup}
We use Qwen-3-Omni \cite{xu2025qwen3omnitechnicalreport} as the base
model used for ADEPT and train on 169K samples using GRPO.
We use an effective batch size of 256 prompts and sample $K{=}8$ rollouts per prompt.
Training runs for 2 epochs (1,250 steps in total), yielding 2.56M trajectories.
We optimize a composite reward that is defined in section \ref{sec:grpo_optimization}.

\paragraph{Baselines.}
We compare the following groups:
(1) \textbf{SSL SER classifiers (primary-only):} HuBERT, wav2vec2.0, and WavLM (frozen / fine-tuned on MSP-Podcast v1.12), which output a single primary label and thus cannot be used to predict minor
emotions.
(2) \textbf{Generative SpeechLLM/MLLMs (primary+secondary):} SALMONN, Qwen-2-Audio (prosody-strong general models), and BLSP-Emo (emotion-specialized), prompted to output both a primary and a minor emotion under a unified schema and decoding constraints.
(3) \textbf{Variations of our models:} Qwen-3-Omni (no tools), ADEPT (w/o GRPO), and ADEPT + GRPO.

\begin{table*}[t]
\caption{
Main results on the MSP-Podcast Test1.
We report strict primary recognition (P-MacroF1: Primary Macro-F1), soft retention of the ground-truth primary label (Soft R: Soft Macro Recall), and set-based recovery of co-occurring emotions (Set R: Set Recall; Jaccard).
Note that standard SSL baselines (e.g., WavLM) are trained for primary emotion classification and cannot output emotion sets.
}
\centering
\footnotesize 
\setlength{\tabcolsep}{3.5pt}  
\begin{tabular}{lccccc}
\toprule
\textbf{Method} 
& \textbf{Avg Size}
& \textbf{P-MacroF1} $\uparrow$
& \textbf{Soft R} $\uparrow$
& \textbf{Set R} $\uparrow$
& \textbf{Jaccard} $\uparrow$ \\
\midrule
\multicolumn{6}{l}{\textit{Reported SSL baselines on MSP-Podcast (primary-only classification)}} \\
HuBERT \cite{Hsu2021HuBERTMaskedPrediction} 
& -- & 0.2850 & -- & -- & -- \\
wav2vec2 \cite{baevski2020wav2vec20frameworkselfsupervised}
& -- & 0.2380 & -- & -- & -- \\
WavLM (Frozen) \cite{Chen2021WavLM} 
& -- & 0.2970 & -- & -- & -- \\
WavLM (Fine-tuned) \cite{naini2025interspeech_ser_challenge} 
& -- & \textbf{0.3640} & -- & -- & -- \\
\midrule
\multicolumn{6}{l}{\textit{Reported SLLM/MLLM baselines (generative, primary+secondary capable)}} \\
SALMONN \cite{Tang2023SALMONN}       
& 1.76 & 0.1389 & 0.4320 & 0.3320 & 0.2280 \\
BLSP-Emo \cite{wang2024blspemo}      
& 2.08 & 0.2915 & 0.6740 & 0.5320 & 0.4280 \\
Qwen-2-Audio \cite{chu2024qwen2audiotechnicalreport}   
& 1.95 & 0.2290 & 0.5120 & 0.4150 & 0.3260 \\
\midrule
\multicolumn{6}{l}{\textit{Our baselines and ADEPT variants (primary+secondary prediction)}} \\
Qwen-3-Omni\cite{xu2025qwen3omnitechnicalreport} 
& 2.02 & 0.2358 & 0.6003 & 0.4923 & 0.3890 \\
ADEPT (w/o GRPO) 
& 2.15 & 0.3571 & 0.7032 & 0.5748 & 0.4450 \\
ADEPT + GRPO 
& 2.21 & \textbf{0.4224} & \textbf{0.7874} & \textbf{0.6192} & \textbf{0.4751} \\
\bottomrule
\end{tabular}
\vspace{-8pt}
\label{tab:main_results}
\end{table*}

\subsection{Evaluation Metrics}
\label{subsec:metrics}

To quantify ADEPT's ability to preserve the consensus label while recovering minority ambiguity, we report metrics at three levels:
(i) primary-label fidelity, (ii) set-level ambiguity decoding, and (iii) ambiguity calibration.
Let $y_{\text{pri}}$ and $\hat{y}_{\text{pri}}$ denote the ground-truth and predicted primary labels, respectively, while $\mathcal{Y}_{\text{gt}}$ and $\hat{\mathcal{Y}}$ represent the full ground-truth and predicted label sets (primary + minor).

\noindent\textbf{Primary-label fidelity.}
We report \textit{Primary Macro-F1} under strict matching ($\hat{y}_{\text{pri}} = y_{\text{pri}}$), which is robust to class imbalance in the MSP-Podcast corpus. To assess primary emotion preservation under ambiguity-aware decoding, we also report \textit{Soft Recall}, defined as relaxed primary-label matching,
$\mathcal{R}_{\text{soft}}=\frac{1}{N}\sum_{i=1}^{N}\mathbb{I}[y_{\text{pri}}^{(i)} \in \hat{\mathcal{Y}}^{(i)}]$.

\vspace{2pt}
\noindent\textbf{Set-level ambiguity decoding.}
We report \textit{Set Recall},
$\mathcal{R}_{\text{set}}=\frac{|\mathcal{Y}_{gt}\cap\hat{\mathcal{Y}}|}{|\mathcal{Y}_{gt}|}$,
to quantify coverage of co-occurring emotions.
To penalize over-prediction, we additionally report the \textit{Jaccard Index (IoU)},
$\mathrm{IoU}=\frac{|\mathcal{Y}_{gt}\cap\hat{\mathcal{Y}}|}{|\mathcal{Y}_{gt}\cup\hat{\mathcal{Y}}|}$.

\vspace{2pt}
\noindent\textbf{Calibration.}
We report \textit{Average Cardinality},
$\frac{1}{N}\sum|\hat{\mathcal{Y}}|$,
to monitor the mean predicted set size (i.e., number of emotional classes predicted) and diagnose whether the model’s ambiguity threshold is overly conservative or permissive relative to the ground-truth annotation density.

\subsection{Main Results on the MSP-Podcast corpus}
\label{subsec:main_results_msp}

Table~\ref{tab:main_results} presents the quantitative comparison on the MSP-Podcast (Test1) set. We evaluate primary-label fidelity using Primary Macro-F1 and Soft Recall, and assess set-level ambiguity decoding using Set Recall and Jaccard.

\textbf{Outperforming supervised SSL baselines.}
Among traditional SSL classifiers, the fine-tuned WavLM achieves the strongest primary-only performance with a Primary Macro-F1 of 0.3640.
While ADEPT (w/o GRPO) already reaches a comparable Primary Macro-F1 (0.3571), ADEPT + GRPO further improves it to 0.4224, outperforming fine-tuned WavLM by an absolute margin of +0.0584 (+5.8 points).

\textbf{Outperforming generative SpeechLLM/MLLM baselines.}
Compared to generative models capable of set prediction (SALMONN, Qwen-2-Audio, and BLSP-Emo), ADEPT consistently achieves higher scores across all metrics.
In particular, ADEPT + GRPO attains the best Soft Macro Recall (0.7874) and Jaccard (0.4751), demonstrating stronger recovery of co-occurring minority emotions.

\textbf{Effect of tool protocol and GRPO.}
Starting from the tool-free Qwen-3-Omni baseline (0.2358 Primary Macro-F1, 0.6003 Soft R), enabling the ADEPT protocol yields substantial gains (0.3571 Primary Macro-F1, 0.7032 Soft R).
GRPO post-training further strengthens both primary and set-based performance while keeping the average prediction size nearly unchanged (2.02 $\rightarrow$ 2.21). These results suggest that improvements come from precise evidence-supported disambiguation rather than indiscriminate label expansion.

\begin{figure}[t]
    \centering
    \includegraphics[width=\linewidth]{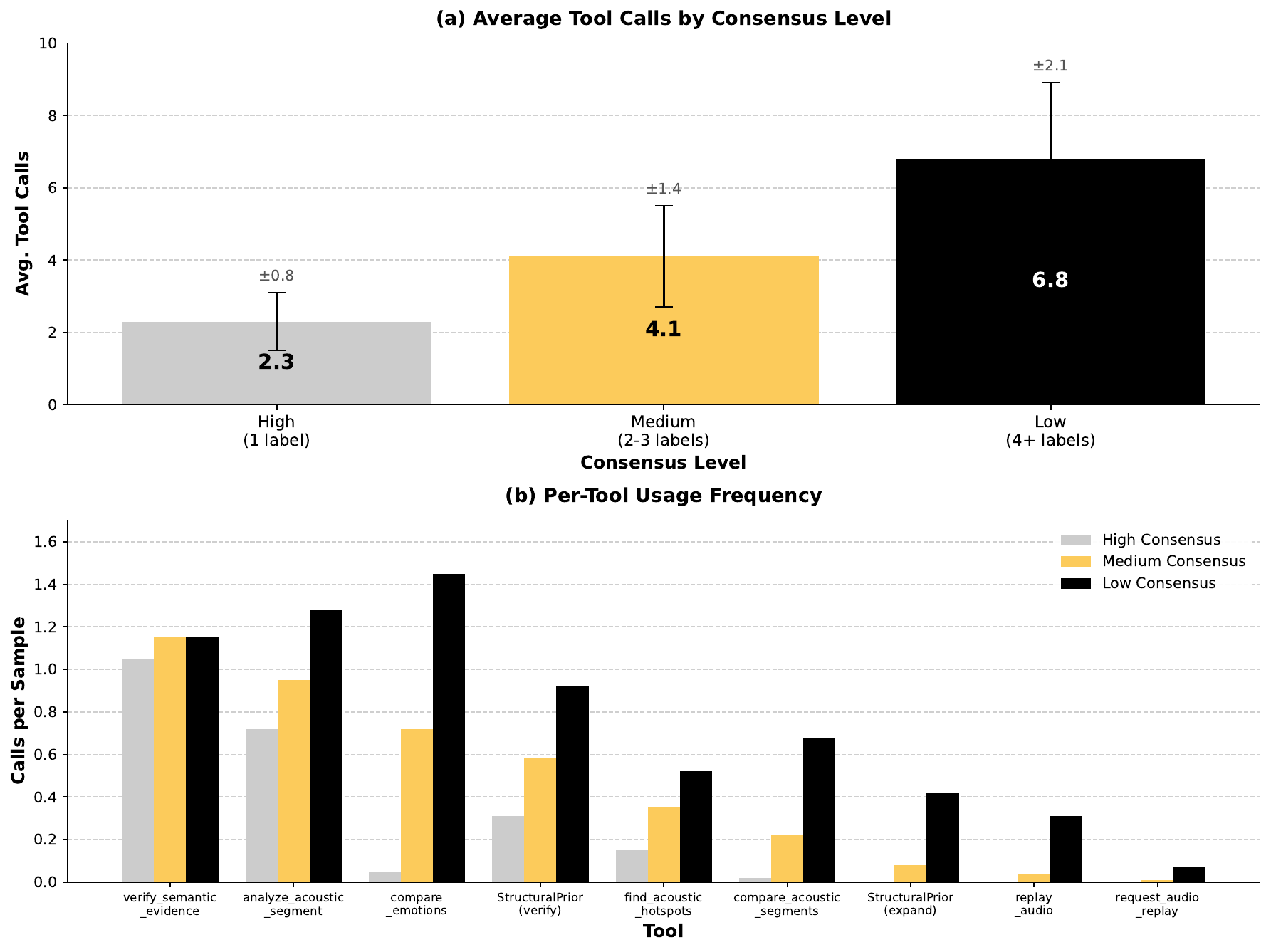}
\caption{\textbf{Tool usage scales with annotation ambiguity.} 
(a) Average tool calls increase from 2.3 (high consensus) to 6.8 (low consensus).
(b) Low-consensus samples show elevated usage across all tools, especially \texttt{replay\_audio} and \texttt{StructuralPrior(expand)}.}
    \label{fig:tool_calls_by_consensus}
\end{figure}


\subsection{Ambiguity-Driven Tool Budget Allocation}
\label{sec:exp_tool_calls_by_consensus}
Figure~4(a) shows the average number of tool calls per sample across three consensus levels, which is measured by the cardinality of the ground-truth emotion set per utterance (including both primary and minor labels): High consensus corresponds to a single label, Medium consensus to 2--3 labels, and Low consensus to $4+$ labels. This figure shows ambiguity-driven budget allocation in ADEPT: the average number of tool calls increases monotonically as annotator consensus decreases (2.3 for High, 4.1 for Medium, and 6.8 for Low consensus).
This result indicates that the agent spends minimal computation on easy samples while reserving more evidence acquisition and refinement steps for intrinsically ambiguous instances. Figure~\ref{fig:tool_calls_by_consensus}(b) further shows that the increase is \emph{tool-specific} rather than uniform: Low-consensus samples trigger substantially more contrastive verification and closed-loop refinement, with pronounced rises in \texttt{compare\_emotions}, \texttt{StructuralPriorTool}, and \texttt{replay\_audio}. Together, these results provide direct evidence that ADEPT implements \textbf{on-demand computation} under uncertainty, allocating tool calls to the most ambiguous utterances instead of uniformly spending budget across the dataset.

\begin{table}[t]
\caption{
\textbf{Ablation results on the MSP-Podcast Test1.}
We ablate the tool protocol and GRPO alignment (top). Then we remove each tool family while keeping other components unchanged (bottom).
}
\centering
\footnotesize
\setlength{\tabcolsep}{2.6pt}
\begin{tabular}{l|cccc}
\toprule
\textbf{Variant} 
& \textbf{Set R} 
& \textbf{Soft R} 
& \textbf{Jaccard} 
& \textbf{P-F1} \\
\midrule
\multicolumn{5}{l}{\textit{Protocol \& alignment ablations}} \\
Omni3 (no tools) 
& 49.23 & 60.03 & 38.90 & 23.58 \\
ADEPT (w/o GRPO) 
& 57.48 & 70.32 & 44.50 & 35.71 \\
ADEPT (Full) 
& \textbf{61.92} & \textbf{78.74} & \textbf{47.51} & \textbf{42.24} \\
\midrule
\multicolumn{5}{l}{\textit{Tool-family ablations}} \\
w/o Prior 
& 58.17 & 74.58 & 44.63 & 39.07 \\
w/o Semantic 
& 60.06 & 75.93 & 42.31 & 33.86 \\
w/o Acoustic 
& 59.08 & 75.21 & 41.28 & 34.52 \\
w/o Replay 
& 60.47 & 77.06 & 46.18 & 40.44 \\
w/o Trust Gate 
& 60.73 & 77.82 & 44.96 & 40.63 \\
\bottomrule
\end{tabular}
\vspace{-2pt}
\label{tab:ablation_tool_family}
\end{table}

\subsection{Ablation Studies}
\label{subsec:ablation_studies}

We conduct ablations to identify the sources of ADEPT's improvements, focusing on (i) the effect of the tool protocol and GRPO alignment, and (ii) the contribution of individual components.
As shown in Table~\ref{tab:ablation_tool_family}, enabling the ADEPT protocol without GRPO already yields substantial gains over the tool-free Omni3 baseline, improving Set Recall from 49.23 to 57.48 and Jaccard from 38.90 to 44.50.
GRPO alignment further strengthens both primary and set-based performance, reaching 61.92 Set Recall and 47.51 Jaccard.

We then ablate one component at a time while keeping the remaining parts unchanged. Table \ref{tab:ablation_tool_family} shows that removing prior tools mainly hurts recall-oriented metrics, indicating their importance for candidate routing and confuser-aware exploration.
Removing semantic tools causes the largest drop in Primary Macro-F1 and a notable decrease in Jaccard, highlighting the role of transcript-based verification for contrastive disambiguation.
Disabling acoustic tools yields the most severe degradation in Jaccard, suggesting that localized perceptual evidence is critical for resolving ambiguity in naturalistic speech.
Replay removal also degrades performance, showing that closed-loop re-injection improves recovery under high uncertainty.
Finally, ablating the trust gate reduces overall set quality, consistent with its role in preventing the policy from over-optimizing tool-related rewards. Overall, ADEPT benefits from the complementary components of the tools rather than a single dominant factor.
Additional sensitivity analyses on the GRPO rollout size $K$ and reward weights are provided in Appendix~\ref{sec:appendix_grpo_sensitivity}.

\begin{table}[t]
\caption{
\textbf{Zero-shot generalization on the \textsc{IEMOCAP} corpus (mapped 7-way).}
We report Acc (accuracy) and set-based metrics (Soft R: Soft Recall; Set R: Set Recall; Jaccard); $^\dagger$ Evaluated on the shared label subset.
}
\centering
\footnotesize
\resizebox{\linewidth}{!}{
\setlength{\tabcolsep}{3.5pt} 
\begin{tabular}{l|cccc}
\toprule
\multirow{2}{*}{\textbf{Method}} & \multicolumn{4}{c}{\textbf{IEMOCAP (Mapped 7-way$^\dagger$)}} \\
\cmidrule(l){2-5}
& \textbf{Acc} $\uparrow$ 
& \textbf{Soft R} $\uparrow$ 
& \textbf{Set R} $\uparrow$ 
& \textbf{Jaccard} $\uparrow$ \\
\midrule
\multicolumn{5}{l}{\textit{Supervised / Fine-tuned baselines}} \\
Emotion-LLaMA
& 55.47 & 63.82 & 38.25 & 34.11 \\
BLSP-Emo
& \textbf{76.13} & \textbf{80.51} & 52.30 & \textbf{48.90} \\
\midrule
\multicolumn{5}{l}{\textit{Zero-shot / Generalist models}} \\
Qwen-2-Audio
& 37.71 & 45.27 & 22.15 & 19.83 \\
Qwen-3-Omni (Baseline)
& 42.37 & 51.12 & 35.24 & 31.08 \\
ADEPT + GRPO (Ours)
& \underline{54.83} & \underline{67.42} & \underline{\textbf{54.80}} & \underline{43.15} \\
\bottomrule
\end{tabular}
}
\vspace{-2pt}
\label{tab:generalization_iemocap}
\end{table}

\subsection{Zero-Shot Generalization on the IEMOCAP corpus}
\label{subsec:generalization_iemocap}

To assess robustness under domain shift, we evaluate ADEPT on the IEMOCAP corpus in a strict zero-shot setting, distinct from baselines fully fine-tuned on the target data. We align the taxonomies for a valid 7-way evaluation (Anger, Sadness, Happiness, Neutral, Fear, Surprise, Disgust) by merging \textit{Excited} into \textit{Happy}, excluding \textit{Frustrated} and \textit{Other} due to the lack of semantic equivalents, and noting that \textit{Contempt} is absent in the IEMOCAP labels. Crucially, to validate our ambiguity-aware modeling, we apply the identical label construction protocol used for the MSP-Podcast corpus. Leveraging the 3-annotator ground truth per utterance, we derive the primary emotion from the plurality vote (majority) and the minor emotion set from remaining non-majority votes (see Appendix~\ref{sec:appendix_iemocap_stats} for detailed distribution statistics). As shown in Table~\ref{tab:generalization_iemocap}, while supervised models (e.g., BLSP-Emo) naturally excel in single-label Accuracy by fitting the model, ADEPT achieves a Set Recall of 54.80\%, effectively surpassing the supervised SOTA (52.30\%).
This inversion highlights a fundamental trade-off: standard classifiers suppress ambiguity to optimize the primary class, whereas ADEPT's evidence-seeking policy successfully retrieves valid co-occurring emotions, even for rare classes such as Disgust in zero-shot.

\section{Conclusion}
\label{sec:conclusion}
We present ADEPT, an agentic framework that reframes Speech Emotion Recognition from static classification into a dynamic, evidence-grounded inquiry process.
By enforcing Explicit Information Retrieval, ADEPT decouples perception from interpretation and resolves emotional ambiguity through structured, theory-driven probing. Experiments on the MSP-Podcast corpus show consistent gains in primary accuracy and the recovery of co-occurring minority emotions.
Strong zero-shot performance on the IEMOCAP corpus further demonstrates robustness to domain shift over latent feature matching. Overall, this work advances auditable Affective Computing by moving from opaque prediction toward verifiable, evidence-based emotion reasoning.

\section*{Impact Statement}

This paper introduces ADEPT, a framework designed to advance Speech Emotion Recognition (SER) from static classification to auditable, evidence-grounded reasoning. Our work has potential societal impacts in three primary areas:

\paragraph{Modeling Emotional Complexity and Inclusivity.}
Standard SER approaches typically enforce consensus by discarding minority annotations as noise. This \emph{winner-takes-all} paradigm risks oversimplifying the complexity of human affect and silencing valid, nuanced emotional expressions. By shifting the paradigm to \emph{Ambiguity-Driven Emotion Reasoning}, our work explicitly validates minority perspectives and recognizes that human emotion is often pluralistic and mixed. This is ethically significant, as it moves AI toward a more human-centric understanding that respects subjective variation and ensures that subtle or complex emotional states, which are often overlooked by majority-voting systems, are preserved and recognized.

\paragraph{Trust and Transparency in High-Stakes Applications.}
As Large Language Models (LLMs) are increasingly integrated into daily life and critical systems, the demand for emotional intelligence must be matched by a demand for explainability. In high-stakes domains such as automated customer service, mental health screening, and empathetic human--computer interaction, an opaque \emph{black-box} decision is insufficient and potentially harmful. ADEPT addresses this challenge by enforcing \emph{Explicit Information Retrieval (EIR)}, ensuring that every emotional judgment is causally linked to verifiable acoustic and semantic evidence. Such transparency is crucial for user safety, as it allows practitioners to audit the system's reasoning process and identify potential hallucinations or biases before they lead to adverse outcomes.

\paragraph{Potential Risks and Mitigation.}
While improved emotional understanding can enhance human--AI interaction and AI companionship, we acknowledge that advanced SER technologies could be misused for intrusive surveillance or emotional manipulation. We argue that the agentic, tool-based nature of our framework offers a degree of mitigation compared to end-to-end latent models. By requiring the system to explicitly expose its audit trail, specifically indicating which audio segments or textual spans support a given conclusion, our approach facilitates human oversight and accountability. This design choice makes it easier to monitor system behavior, diagnose ethical failures, and supports more responsible deployment of affective computing technologies.


\bibliography{example_paper}
\bibliographystyle{icml2026}

\newpage
\appendix
\onecolumn

\section{Appendix: MSP-Podcast Corpus and Annotation Details}
\label{sec:appendix_msp}

\paragraph{Dataset Overview and Construction.}
We utilize the MSP-Podcast corpus (v2.0) \cite{Busso2025MSPPodcast}, a large-scale dataset of naturalistic emotional speech collected from online podcast recordings. 
Unlike acted emotion corpora, the MSP-Podcast corpus is constructed via a machine learning-driven retrieval pipeline designed to mitigate data sparsity in spontaneous emotional expression \cite{Lotfian_2019_3}. Specifically, the pipeline automatically segments long-form podcast audio and leverages pretrained emotion recognition models to identify and retain segments enriched with affective content, improving coverage across the emotional spectrum.
The final release contains \textbf{267,905 speaking turns} (approximately \textbf{409 hours}) from at least \textbf{3,641 unique speakers}, making it one of the largest publicly available naturalistic speech emotion resources. Detailed split is showed in table \ref{tab:msp_splits}.

\begin{table}[t]
\centering
\small
\setlength{\tabcolsep}{6pt}
\caption{Official partitions and speaker counts in the MSP-Podcast corpus.}
\label{tab:msp_splits}
\begin{tabular}{lrr}
\hline
\textbf{Split} & \textbf{\# Turns} & \textbf{\# Speakers} \\
\hline
Train & 169,190 & 2,220 \\
Dev   & 34,399  & 704 \\
Test1 & 46,294  & 465 \\
Test2 & 14,822  & 112 \\
Test3 & 3,200   & 428 \\
\hline
All   & 267,905 & 3,641 \\
\hline
\end{tabular}
\end{table}

\begin{table}[]
\caption{\textbf{Annotation ambiguity statistics of the MSP-Podcast corpus (v2.0).} 
We report tie prevalence and the distribution of label multiplicity (primary + minor) for Train and Test1 splits, highlighting the multi-label and ambiguity-rich nature of naturalistic emotional speech.}
\centering
\small
\setlength{\tabcolsep}{5pt}
\begin{tabular}{lcc}
\toprule
\textbf{Metric} & \textbf{Train} & \textbf{Test1} \\
\midrule
Total samples & 169,087 & 46,250 \\
Tie count & 30,096 & 8,509 \\
Tie rate & 17.8\% & 18.4\% \\
Avg labels per audio & 2.55 ($\pm$1.00) & 2.58 ($\pm$1.10) \\
Median labels & 3.0 & 3.0 \\
Min labels & 1 & 1 \\
Max labels & 8 & 8 \\
Avg primary labels & 1.24 & 1.28 \\
Avg minor labels & 1.31 & 1.30 \\
\bottomrule
\end{tabular}
\label{tab:msp_annotation_stats}
\end{table}

\paragraph{Categorical Emotion Annotation Protocol.}
To ensure annotation reliability, each speaking turn is labeled through a perceptual evaluation process with a minimum of five raters per segment. The early recordings
were annotated with a modified version of the crowdsourcing protocol presented in  \cite{Burmania_2016_2}. The remaining 65.8\% of the annotations were conducted by trained student workers (see details in \cite{Busso2025MSPPodcast}). The annotation interface is designed to capture the nuanced and often mixed nature of emotions in natural conversational speech through a two-stage procedure.
Concretely, the survey presents two consecutive categorical questions:

\begin{enumerate}[leftmargin=*, itemsep=2pt, topsep=2pt]
    \item \textbf{Primary Emotion (Single-Label).}
    Annotators first identify the dominant emotion by answering: \emph{``Is any of these emotions the primary emotion in the audio?''}
    They select exactly one label from eight standard categories: \textit{Anger, Sadness, Happiness, Surprise, Fear, Disgust, Contempt,} and \textit{Neutral} (or \textit{Other}).
    The consensus primary label is aggregated using the plurality rule.

    \item \textbf{Secondary Emotions (Multi-Label).}
    To account for mixed affect (e.g., \textit{happily surprised}), annotators are subsequently asked to select \emph{all} perceived emotional classes.
    This stage expands the taxonomy to a list of 16 emotions set:
    \textit{Anger, Frustration, Disgust, Annoyance, Sadness, Depression, Disappointment, Fear, Happiness, Surprise, Excitement, Contempt, Amusement, Concern, Confusion,} and \textit{Neutral} (plus \textit{Other}).
\end{enumerate}

\paragraph{Scope in This Work (8-Way Primary Labels Only).}
In this study, we only use the \textbf{first-stage categorical annotations} (the 8-way primary emotion taxonomy) to maintain a closed-set label space and enable consistent evaluation under standard SER settings.
We intentionally do not use the second-stage expanded labels, as they introduce a taxonomy mismatch and additional label granularity that is not uniformly comparable to prior SER baselines. Moreover, the first-stage labels are the most widely adopted and reliably aggregated setting (via plurality voting), which allows us to isolate ADEPT’s contribution in terms of explicit evidence retrieval and tool-grounded verification, without confounding factors from heterogeneous secondary label definitions. We leave the integration of expanded secondary labels and richer affect taxonomies as a promising direction for future work.

\begin{figure}[t]
    \centering
    \includegraphics[width=\linewidth]{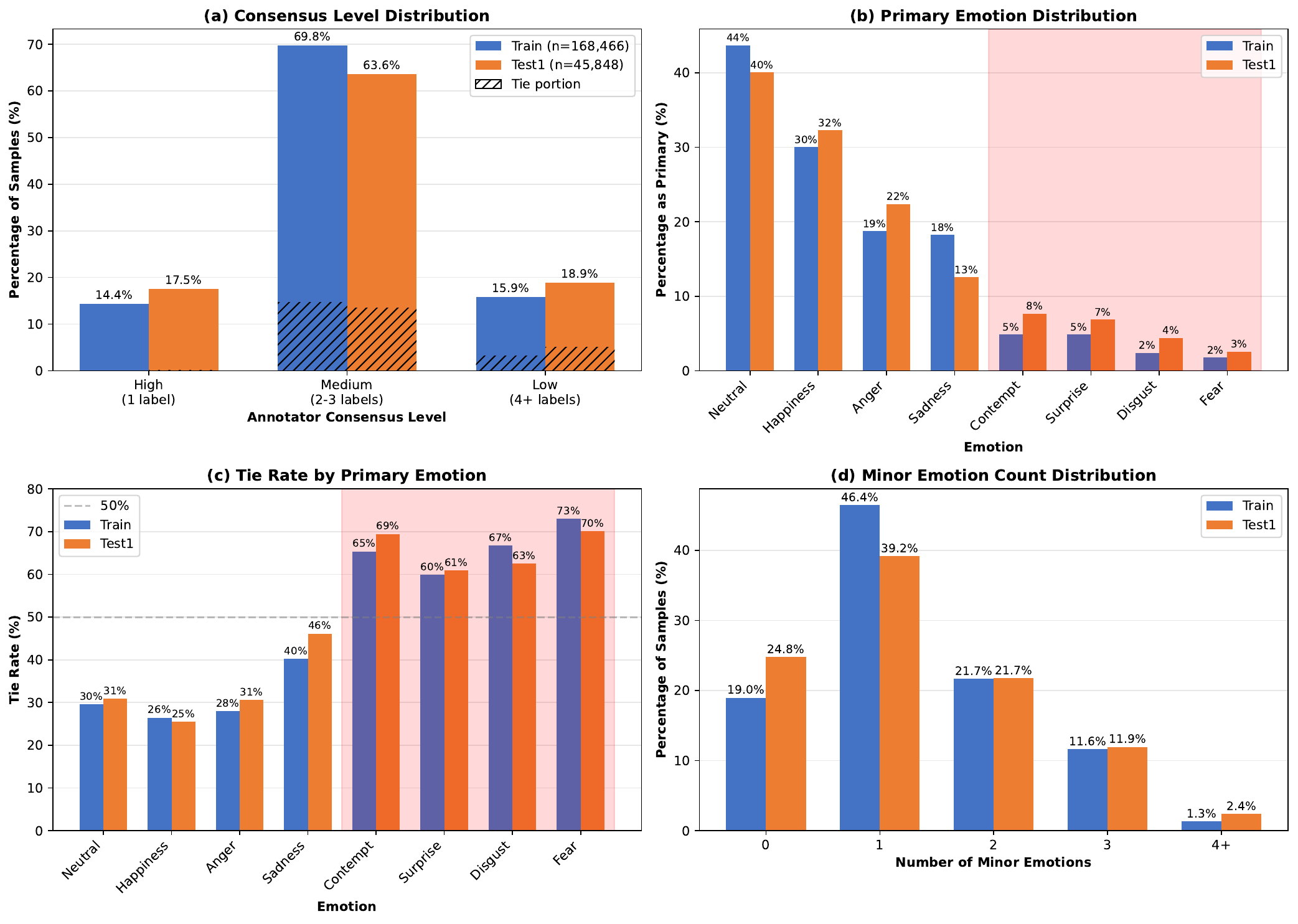}
    \caption{\textbf{Dataset overview of the MSP-Podcast corpus (v2.0) and annotation ambiguity.} 
    We summarize (a) the annotator consensus level distribution (high: 1 label; medium: 2--3 labels; low: 4+ labels), 
    (b) the primary emotion distribution, 
    (c) the tie rate conditioned on the primary emotion, and 
    (d) the distribution of minor emotion counts for both Train and Test1 splits. 
    These statistics highlight the prevalence of disagreement and multi-label ambiguity in naturalistic emotional speech, motivating ambiguity-driven emotion reasoning.}
    \label{fig:msp_dataset_overview}
\end{figure}

\paragraph{Annotation Distribution and Ambiguity Statistics.}
To better characterize the annotation structure of the MSP-Podcast corpus, we summarize key distributional properties in Figure~\ref{fig:msp_dataset_overview} and Table~\ref{tab:msp_annotation_stats}. 
Figure~\ref{fig:msp_dataset_overview}(a) reports the \emph{consensus level distribution}, where the majority of samples fall into the medium-consensus regime (2--3 labels), while a non-trivial portion exhibits low consensus (4+ labels), reflecting substantial ambiguity in naturalistic emotional speech.
Figure~\ref{fig:msp_dataset_overview}(b) shows the \emph{primary emotion distribution}, which is highly imbalanced: a small subset of frequent categories dominates the corpus, whereas less common emotions (e.g., \textit{Contempt}, \textit{Surprise}, \textit{Disgust}, \textit{Fear}) occur sparsely as primary labels.
Figure~\ref{fig:msp_dataset_overview}(c) further reveals that these peripheral emotions tend to have markedly higher tie rates, suggesting that they are both harder to disambiguate and more likely to co-occur with semantically or acoustically neighboring categories.
Finally, Figure~\ref{fig:msp_dataset_overview}(d) summarizes the \emph{minor emotion count distribution}. 
We observe that most utterances contain at least one minor emotion (Train: 81\%; Test1: 76\%), and the most common case is exactly one minor label (Train: 46\%; Test1: 39\%), indicating that multi-label structure is prevalent rather than an exception.

Across splits, tie cases account for approximately one-fifth of the corpus (Train: 17.8\%; Test1: 18.4\%), and each audio segment contains on average 2.55--2.58 emotion labels (primary + minor), with a median of 3 labels.
These statistics confirm that MSP-Podcast is not only class-imbalanced but also intrinsically ambiguity-rich, motivating our ambiguity-driven emotion reasoning setting where both minority votes and tie structures are treated as informative supervision signals rather than annotation noise.
Notably, tie cases correspond to \emph{no-agreement} instances under a single-consensus labeling regime (i.e., no unique majority emotion emerges from annotator votes). 
In many prior SER pipelines, such ambiguous samples are routinely discarded to enforce a single-label training objective, leading to substantial data waste and systematically removing precisely those utterances that reflect the multiplicity and co-occurrence of human affect.
In particular, the elevated tie rates and higher minor-label multiplicity in peripheral categories indicate that ambiguous samples require additional evidence acquisition, aligning with ADEPT's design principle of allocating more tool-use budget to high-uncertainty cases.

\begin{figure}[t]
    \centering
    \includegraphics[width=0.6\linewidth]{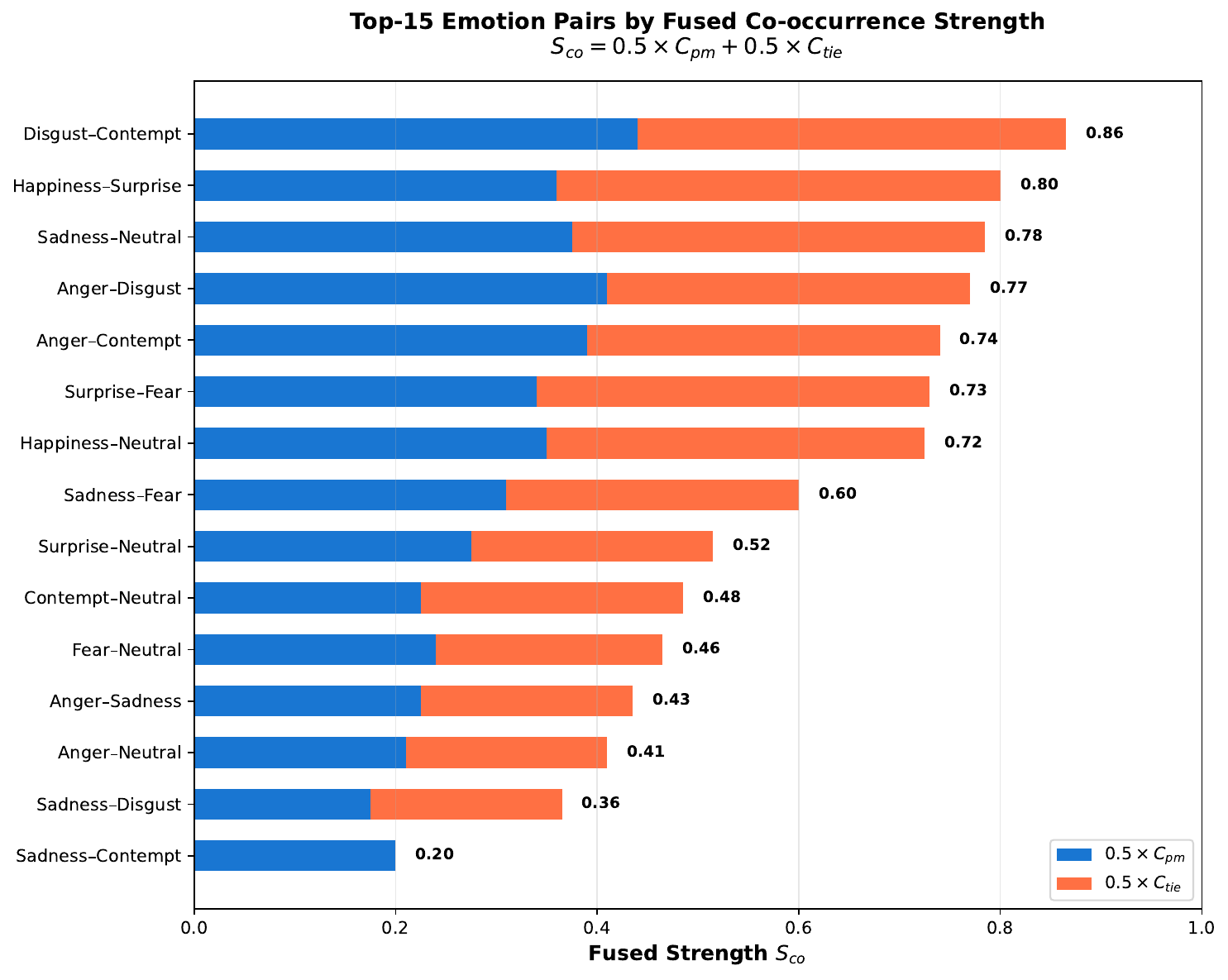}
    \caption{
    Top emotion pairs ranked by fused co-occurrence strength $S_{\mathrm{co}}$, computed as
    $S_{\mathrm{co}}(a,b)=0.5\,\mathbf{C}^{\mathrm{pm}}(a,b)+0.5\,\mathbf{C}^{\mathrm{tie}}(a,b)$.
    This ranking provides a baseline-free structural prior for Phase-2 budget allocation by prioritizing high-coupling candidate pairs for contrastive verification under the EIR constraint.
    }
    \label{fig:fused_top_pairs}
\end{figure}

\section{Appendix : Structural Prior Tool from Co-occurrence Statistics}
\label{sec:appendix_prior}

\paragraph{Motivation.}
Phase-1 in ADEPT intentionally performs a high-recall initialization to preserve ambiguity under annotator disagreement.
As a result, the Phase-2 agent must allocate a limited probing budget to the most informative comparisons among candidate emotions, and occasionally recover from Phase-1 recall bottlenecks via backtracking.
To support this process without introducing model-dependent biases, we define a lightweight \emph{Structural Prior Function} derived solely from corpus-level \emph{emotion co-occurrence statistics}.
Importantly, this prior does not predict emotions, does not prescribe tool usage, and is never treated as admissible evidence for final adjudication.
Instead, it provides \emph{pair-level scheduling signals} that help the agent prioritize which candidate pairs to verify under the EIR constraint.

\subsection{Emotion Label Space and Outputs}
We consider a fixed 8-way categorical label space:
\[
\mathcal{Y}=\{\textsc{Anger}, \textsc{Sadness}, \textsc{Happiness}, \textsc{Surprise}, 
\textsc{Fear}, \textsc{Disgust}, \textsc{Contempt}, \textsc{Neutral}\}.
\]

In addition to predicting the primary emotion, ADEPT explicitly outputs a \emph{tie prediction} flag indicating whether the instance exhibits a high-uncertainty tie between two competing emotions, along with the predicted tie pair when applicable.

\begin{figure}[t]
    \centering
    \includegraphics[width=\linewidth]{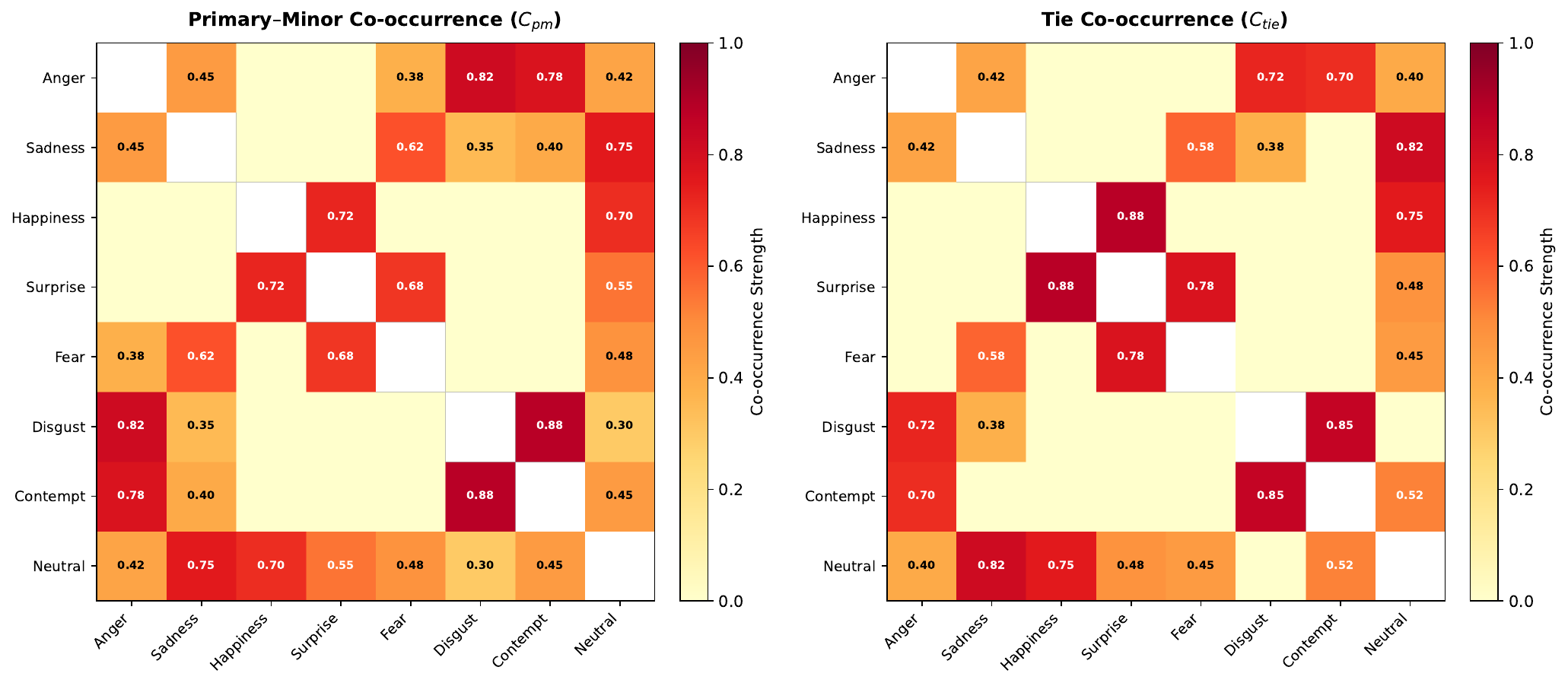}
    \caption{\textbf{Co-occurrence structures induced by annotator disagreement.}
    Left: normalized \emph{primary--minor} co-occurrence matrix $\tilde{\mathbf{C}}^{\mathrm{pm}}$, capturing systematic minor emotions that co-exist with perceived primary affect (multiplicity).
    Right: normalized \emph{tie} co-occurrence matrix $\tilde{\mathbf{C}}^{\mathrm{tie}}$, characterizing high-confusion emotion pairs that frequently emerge as top-vote ties.
    Both matrices are computed over the 8-way label space and provide pair-level scheduling signals for Phase-2 verification, rather than admissible evidence for final adjudication.}
    \label{fig:cooccur_heatmaps}
\end{figure}

\subsection{Co-occurrence Structures from Annotator Disagreement}
We construct two complementary co-occurrence structures that capture distinct sources of ambiguity in human perception.

\paragraph{(i) Primary--Minor Co-occurrence.}
Given an utterance $n$, let $\mathcal{P}_n$ denote its primary emotion set (allowing ties) and $\mathcal{M}_n$ denote its minor emotion set (minority-vote emotions).
We define a \emph{primary--minor} co-occurrence count matrix $\mathbf{C}^{\mathrm{pm}} \in \mathbb{R}^{|\mathcal{Y}|\times|\mathcal{Y}|}$ by accumulating:
\begin{equation}
\mathbf{C}^{\mathrm{pm}}[p,m] \;{+}{=}\; 1 \quad \forall p \in \mathcal{P}_n,\; m \in \mathcal{M}_n.
\label{eq:cooccur_pm}
\end{equation}
This structure reflects \emph{Complexity}: secondary emotions that systematically co-exist with the perceived primary affect.

\paragraph{(ii) Tie Co-occurrence.}
To model \emph{systematic perceptual ties}, we additionally construct a tie count matrix $\mathbf{C}^{\mathrm{tie}}$.
For any utterance with a top-vote tie $\mathcal{P}_n=\{a,b\}$, we accumulate:
\begin{equation}
\mathbf{C}^{\mathrm{tie}}[a,b] \;{+}{=}\; 1,\quad
\mathbf{C}^{\mathrm{tie}}[b,a] \;{+}{=}\; 1.
\label{eq:cooccur_tie}
\end{equation}
Unlike $\mathbf{C}^{\mathrm{pm}}$, which captures primary--secondary mixtures, $\mathbf{C}^{\mathrm{tie}}$ explicitly characterizes \emph{high-confusion emotion pairs} that frequently arise as unresolved alternatives under annotator disagreement. Figure~\ref{fig:cooccur_heatmaps} visualizes both co-occurrence structures.

\begin{figure}[t]
    \centering
    \includegraphics[width=0.5\linewidth]{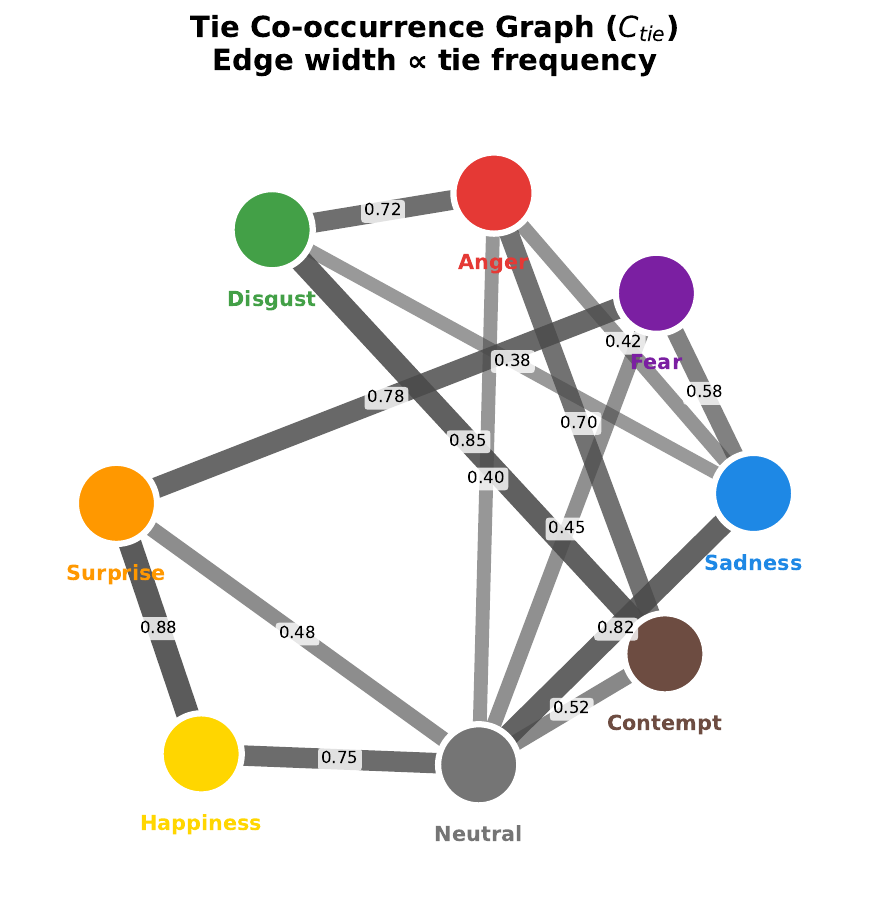}
    \caption{\textbf{Graph view of tie co-occurrence structure.}
    We visualize the normalized tie co-occurrence matrix $\tilde{\mathbf{C}}^{\mathrm{tie}}$ as an undirected graph over the 8 emotion categories, where nodes denote emotions and edge width indicates the frequency of top-vote ties under annotator disagreement.
    This structure highlights high-confusion pairs and provides a tie-oriented scheduling signal for contrastive verification in Phase-2, rather than evidence for instance-level adjudication.}
    \label{fig:tie_graph}
\end{figure}

\subsection{Normalization and Pairwise Prior Strength}
Raw co-occurrence counts are affected by class frequency imbalance.
To obtain a robust pairwise prior signal, we apply symmetric normalization:
\begin{equation}
\tilde{\mathbf{C}}[a,b] \;=\;
\frac{\mathbf{C}[a,b]}
{\sqrt{\mathbf{C}[a,\cdot]\;\mathbf{C}[b,\cdot]}+\epsilon},
\label{eq:cooccur_norm}
\end{equation}
where $\mathbf{C}[a,\cdot]=\sum_{k}\mathbf{C}[a,k]$ and $\epsilon$ is a small constant.
We compute $\tilde{\mathbf{C}}^{\mathrm{pm}}$ and $\tilde{\mathbf{C}}^{\mathrm{tie}}$ separately, and define a fused co-occurrence prior strength:
\begin{equation}
S_{\mathrm{co}}(a,b) \;=\;
\lambda \cdot \tilde{\mathbf{C}}^{\mathrm{pm}}[a,b]
+ (1-\lambda)\cdot \tilde{\mathbf{C}}^{\mathrm{tie}}[a,b],
\label{eq:cooccur_fused}
\end{equation}
where $\lambda \in [0,1]$ is fixed (we use $\lambda=0.5$) to avoid overfitting or tuning to a specific evaluation set. All co-occurrence statistics are computed exclusively on the training split, ensuring that the test set is not used to define these priors.
Crucially, $S_{\mathrm{co}}(a,b)$ is used \emph{only} for ranking candidate pairs and never interpreted as evidence about an individual instance. Figure~\ref{fig:fused_top_pairs} shows the resulting pair ranking induced by $S_{\mathrm{co}}$.

\subsection{StructuralPriorTool: Pair Scheduling and Backtracking}
At inference step $t$, the agent maintains an evolving candidate set $C_t \subseteq \mathcal{Y}$.
The StructuralPriorTool is a callable prior operator that takes $(C_t,\texttt{intent})$ as input, with an optional \emph{anchor} emotion $e^\star \in C_t$ representing the agent's current focus.
The anchor is not a prediction target and is chosen by the agent to condition pair scheduling on its current verification trajectory.

\paragraph{Mode A: Budget Allocation (\texttt{intent=verify}).}
The tool returns a ranked list of \emph{priority pairs} within $C_t$:
\begin{equation}
\texttt{priority\_pairs} \;=\;
\mathrm{TopK}\big\{(a,b)\,:\,a,b\in C_t,\; a<b\big\}
\;\;\text{by}\;\; S_{\mathrm{co}}(a,b).
\label{eq:verify_pairs}
\end{equation}
If an anchor $e^\star$ is provided, the tool prioritizes pairs that include the anchor (e.g., $(e^\star,b)$), then fills the remaining slots from the global ranking within $C_t$.
This output serves as a \emph{budget allocation signal}: it indicates which competing hypotheses should be contrastively verified first, without prescribing how the verification is performed.

\paragraph{Mode B: Backtracking Expansion (\texttt{intent=expand}).}
When the agent triggers backtracking (e.g., evidence conflicts or all current candidates appear implausible), the tool proposes a minimal candidate expansion set:
\begin{equation}
\texttt{suggested\_candidates} \;=\;
\mathrm{TopL}\Big\{c \in \mathcal{Y}\setminus C_t\Big\}
\;\;\text{by}\;\;
\max_{a\in C_t} S_{\mathrm{co}}(a,c).
\label{eq:expand_candidates}
\end{equation}
This provides a recovery mechanism against Phase-1 recall bottlenecks while avoiding uncontrolled candidate explosion.

\paragraph{Tie Pair Prioritization.}
Since ADEPT additionally predicts whether a tie is present, we expose a tie-oriented scheduling signal based on $\tilde{\mathbf{C}}^{\mathrm{tie}}$:
\begin{equation}
\texttt{tie\_priority\_pairs} \;=\;
\mathrm{TopK}\big\{(a,b)\,:\,a,b\in C_t,\; a<b\big\}
\;\;\text{by}\;\; \tilde{\mathbf{C}}^{\mathrm{tie}}[a,b].
\label{eq:tie_pairs}
\end{equation}
This does not assert that an instance is tied; rather, it highlights candidate pairs that are historically tied under annotator disagreement and therefore merit careful contrastive verification. Figure~\ref{fig:tie_graph} provides a graph view of the tie structure.

\subsection{Usage Constraints (Preventing Confirmation Bias)}
To prevent confirmation bias and decision leakage, the StructuralPriorTool is constrained as follows:
(i) it never outputs class probabilities or instance-level label confidence;
(ii) it never emits emotion decisions;
(iii) it never recommends specific tools or fixed execution orders;
(iv) its outputs are used only for Phase-2 scheduling and backtracking, and are excluded from Phase-3 evidence-closed adjudication.
Accordingly, the tool functions as a lightweight \emph{probabilistic context provider} that improves exploration efficiency while preserving evidence-driven integrity under EIR.

\section{Appendix: Semantic Tools}
\label{sec:appendix_semantic}

\paragraph{Theoretical Foundation: Structure Alignment.}
Our semantic verification framework is grounded in the \textbf{Component Process Model (CPM)} \cite{Scherer2001Appraisal}.
Importantly, we adopt the \textbf{7-factor appraisal schema} validated by \citet{Tak2025AwareYetBiased}, a recent study that explicitly operationalizes CPM-style appraisal dimensions for evaluating emotional reasoning in large language models. The paper demonstrates that advanced LLMs (e.g., GPT-4) exhibit measurable structure alignment with human appraisal processes, mapping appraisal factors (e.g., \textit{Suddenness}, \textit{Norm Violation}) to emotion families in a theory-consistent manner.
However, they also report systematic weaknesses on specific dimensions such as \textit{Control/Power}, where models often produce unreliable judgments when the input lacks concrete, text-grounded evidence.

\paragraph{Adaptation to Short Segments (MSP-Podcast).}
Given that the MSP-Podcast corpus segments are often isolated short utterances without conversational history, relying on high-level contextual inference presents a critical challenge.
\citet{Tak2025AwareYetBiased} demonstrated that while LLMs align with appraisal theory structurally, they exhibit high variance and hallucination (e.g., random guesses on \textit{Control/Power}) when contextual evidence is missing.
To mitigate this problem, we operationalize the seven factors \cite{Tak2025AwareYetBiased} into a surface-level linguistic protocol, adopting a high-precision, literal-first strategy.
We transform abstract psychological checks into searches for concrete linguistic markers (e.g., mapping \textit{Urgency} to time adverbs, or \textit{Action Tendency} to fight/flight verbs).
This design deliberately restricts the tool to verify \emph{explicit linguistic realizations} ensuring the agent extracts verbatim evidence spans rather than hallucinating latent context.
Crucially, this strict grounding does not discard holistic emotionality; instead, it shifts the burden of resolving implicit or context-dependent meaning (e.g., sarcasm, ambiguity) to the Acoustic Tools.
In our architecture, the lack of literal semantic evidence serves as a reliable signal to upweight prosodic features, ensuring that ``holistic'' interpretation is grounded in physical acoustic signals rather than textual speculation.

\subsection{The 7-Factor Semantic Protocol}
\label{sec:appendix_semantic_protocol}

Table~\ref{tab:7_factor_mapping} details the mapping between \citet{Tak2025AwareYetBiased} 's appraisal factors and our linguistic search tasks.

\begin{table}[t]
\centering
\caption{\textbf{The 7-factor protocol} adapted from \citet{Tak2025AwareYetBiased} for short-text semantic verification.}
\label{tab:7_factor_mapping}
\resizebox{\columnwidth}{!}{%
\begin{tabular}{l|l|l}
\hline
\textbf{Appraisal Factor (Tak et al., 2025)} &
\textbf{Semantic Verification Task} &
\textbf{Linguistic Cues (Verbatim Spans)} \\
\hline
\textbf{1. Famil\_Sudd} &
Check for \textbf{Novelty / Surprise} &
\textit{Suddenly, wow, what?, really?, unbelievable} \\
\hline
\textbf{2. Neg\_PosConseq} &
Check for \textbf{Valence Polarity} &
\textit{Great, love} (pos) vs.\ \textit{terrible, no, hate} (neg) \\
\hline
\textbf{3. OthSelf\_Causation} &
Check for \textbf{Agency / Blame} &
Pronouns: \textit{you/they} (external) vs.\ \textit{I/me} (internal) \\
\hline
\textbf{4. LoHi\_CoPow} &
Check for \textbf{Control / Dominance} &
Modals: \textit{must, should} (high) vs.\ \textit{can't, help} (low) \\
\hline
\textbf{5. Moral\_Unfair} &
Check for \textbf{Norm Violation} &
\textit{unfair, wrong, liar, cheating, right/wrong} \\
\hline
\textbf{6. Urgency} &
Check for \textbf{Time Pressure} &
\textit{now, immediately, hurry, wait} \\
\hline
\textbf{7. With\_FightAct} &
Check for \textbf{Action Tendency} &
\textit{stop, hit, yell} (attack) vs.\ \textit{leave, hide} (withdraw) \\
\hline
\end{tabular}%
}
\end{table}

\subsection{Tool 1: \texttt{verify\_semantic\_evidence}}
\label{sec:appendix_tool_verify}

This tool performs \textbf{hypothesis validation}. It selects the most discriminative factors from the 7-factor schema for a specific emotion and attempts to locate supporting spans.
Crucially, if the text is too short to contain markers for \textit{LoHi\_CoPow} or \textit{Moral\_Unfair}, the tool returns \texttt{insufficient\_evidence}, routing the system to acoustic probing.

\begin{lstlisting}[
caption={Schema for \texttt{verify\_semantic\_evidence} targeting \textit{Anger}. The tool converts \citet{Tak2025AwareYetBiased}'s factors (e.g., \texttt{Moral\_Unfair}, \texttt{LoHi\_CoPow}) into explicit linguistic searches.},
label={lst:verify_anger},
language=json,
basicstyle=\ttfamily\footnotesize,
breaklines=true
]
{
  "tool": "verify_semantic_evidence",
  "target_emotion": "Anger",
  "appraisal_checks": [
    {
      "factor": "OthSelf_Causation",
      "task": "Identify external blame targets (pronouns/names)."
    },
    {
      "factor": "Moral_Unfair",
      "task": "Extract keywords indicating violation of norms or unfairness."
    },
    {
      "factor": "LoHi_CoPow",
      "task": "Extract imperatives or strong modals indicating high control."
    }
  ],
  "short_text_handling": "If no functional words (pronouns/modals) exist, return evidence=null."
}
\end{lstlisting}

\subsection{Tool 2: \texttt{compare\_emotions}}
\label{sec:appendix_tool_compare}

This tool performs \textbf{disambiguation} between confusing pairs. It uses the 7-factor schema to identify the dimension(s) where two emotions diverge.
For example, distinguishing \textbf{Fear} from \textbf{Sadness} (both negative and low power) relies on \textit{Urgency} and \textit{Action Tendency}.

\begin{lstlisting}[
caption={Schema for compare\_emotions (Fear vs. Sadness). The tool uses Urgency and With\_FightAct for disambiguation \cite{Tak2025AwareYetBiased}.},
label={lst:compare_fear_sad},
basicstyle=\ttfamily\footnotesize,
breaklines=true
]
{
  "tool": "compare_emotions",
  "pair": ["Fear", "Sadness"],
  "divergence_checks": [
    {
      "factor": "Urgency",
      "hypothesis": "Fear implies high urgency (future threat); Sadness implies low urgency (past loss)."
    },
    {
      "factor": "With_FightAct",
      "hypothesis": "Fear implies withdrawal/escape; Sadness implies inaction/resignation."
    }
  ],
  "ambiguity_logic": {
    "trigger": "polysemous_interjection",
    "example": "'Oh no' (could be immediate fear or resigned sadness)",
    "action": "redirect_to_acoustic_probing"
  }
}
\end{lstlisting}

\section{Appendix: Acoustic Tools}
\label{sec:appendix_acoustic_tools}

This appendix details the implementation of our Acoustic Probing Tools, which operationalize Acoustic Explicit Information Retrieval (Acoustic-EIR) as an auditable perceptual interface.
All acoustic tools are \textbf{emotion-neutral}: they only report physical signal properties and are prohibited from emitting emotion labels, confidence scores, or psychological interpretations.

\subsection{Engine III: Acoustic Explicit Information Retrieval (Acoustic-EIR)}
\label{sec:appendix_acoustic_eir}

The Acoustic-EIR engine follows a Dynamic Perception Strategy that decomposes evidence acquisition into three interpretable primitives:
\textbf{Navigation} (where to look), \textbf{Measurement} (what to measure), and \textbf{Comparison} (how evidence differs across regions).
This design avoids the brittleness of static feature extraction that summarizes an entire utterance with a fixed feature set, and instead enables hypothesis-driven, localized probing conditioned on the evolving candidate set.

\subsection{Alignment-based Anchoring (Span $\rightarrow$ Segment)}
\label{sec:appendix_acoustic_alignment}

To ensure acoustic measurements are isomorphic to the agent's linguistic focus, ADEPT leverages pre-computed forced alignment files (e.g., MFA~\citep{mcauliffe17_interspeech}) to map transcript spans to time segments.
Given a transcript span $s=[w_i, \dots, w_j]$, the alignment utility returns a deterministic time interval $[t_s, t_e]$ by aggregating the word-level timestamps.
When spans are discontinuous or partially aligned, we apply a minimal expansion rule by taking the convex hull of aligned words and extending by a small padding window (e.g., $\pm 50$--$100$ ms) to account for boundary uncertainty.
This anchoring guarantees that acoustic evidence is retrieved from the same semantic focus used by the agent in Phase~2, rather than from arbitrary windowing.

\subsection{Tool 1: Navigation via Information-Rich Hotspots}
\label{sec:appendix_acoustic_hotspots}

\paragraph{\texttt{find\_acoustic\_hotspots(audio\_path, focus\_type)}.}
This tool localizes information-rich regions of interest (ROIs) that are likely to contain discriminative acoustic events.
It supports efficient budget allocation by proposing a small set of candidate segments for subsequent measurement.

\begin{itemize}[leftmargin=*]
    \item \textbf{Inputs:} \texttt{audio\_path}; \texttt{focus\_type} $\in$ \{\texttt{energy\_burst}, \texttt{pitch\_excursion}, \texttt{pause\_contrast}, \texttt{voicing\_instability}\}.
    \item \textbf{Outputs:} a ranked list of ROIs $\{[t_s, t_e]\}$, each annotated with a short rationale (e.g., ``energy spike'' or ``rapid F0 change'').
    \item \textbf{Selection rule:} ROIs are ranked by within-segment variance or event magnitude under the specified focus type, with non-maximum suppression to avoid redundant overlaps.
    \item \textbf{Architectural role:} provides \emph{where-to-look} guidance for Acoustic-EIR and reduces unnecessary full-utterance probing.
\end{itemize}

\subsection{Tool 2: Auditable Measurement with Dual-Reference Bucketing}
\label{sec:appendix_acoustic_measurement}

\paragraph{\texttt{analyze\_acoustic\_segment(audio, start, end, metrics)}.}
This tool performs localized acoustic measurements on a target segment and returns \textbf{interpretable evidence bins} rather than raw scalars, bridging the Numeric--Semantic Gap.
Each metric is computed from frame-level signals within $[t_s,t_e]$ and then discretized using a Dual-Reference Bucketing mechanism.

\begin{itemize}[leftmargin=*]
    \item \textbf{Inputs:} segment boundaries \texttt{start}, \texttt{end}; requested \texttt{metrics} list.
    \item \textbf{Raw outputs:} per-metric scalar summaries (e.g., median, IQR, 75th percentile) computed from frame-level measurements.
    \item \textbf{Binned outputs:} categorical evidence descriptors such as \texttt{Low/Mid/High}, \texttt{Stable/Volatile}, and \texttt{Sudden Spike}.
    \item \textbf{Provenance:} each output includes segment timestamps and the normalization reference used, enabling external auditing.
\end{itemize}

\paragraph{Dual-Reference Bucketing.}
For each metric $m(x; [t_s,t_e])$, we compute two normalized scores:
a \textbf{global-reference score} and a \textbf{local-reference score}.
The global reference captures absolute intensity relative to a population baseline (speaker- or dataset-level), while the local reference captures transient deviation relative to the utterance baseline.

\begin{itemize}[leftmargin=*]
    \item \textbf{Global reference (absolute level).} 
    We normalize the segment statistic against a global distribution using robust scaling:
    \[
    z_g = \frac{m - \mathrm{median}(m_{\text{train}})}{\mathrm{IQR}(m_{\text{train}})}.
    \]
    \item \textbf{Local reference (relative deviation).}
    We normalize against the utterance-level baseline to detect emphasis-like events:
    \[
    z_\ell = \frac{m - \mathrm{median}(m_{\text{utt}})}{\mathrm{IQR}(m_{\text{utt}})}.
    \]
\end{itemize}

We then map $(z_g, z_\ell)$ to discrete evidence bins using fixed thresholds (e.g., $\{-1, 0, +1\}$ for \texttt{Low/Mid/High} and a volatility threshold for \texttt{Stable/Volatile}).
This yields descriptors that are interpretable to the agent while remaining traceable to explicit signal measurements.

\subsection{Tool 3: Relational Evidence via Segment Comparison}
\label{sec:appendix_acoustic_comparison}

\paragraph{\texttt{compare\_acoustic\_segments(audio, segments, metrics)}.}
This tool provides contrastive acoustic evidence across multiple segments, which is often more discriminative than absolute measurements.
Given segments $\{[t_s^k,t_e^k]\}_{k=1}^K$, it returns relational statements such as ``Segment A has higher energy burstiness than Segment B''.

\begin{itemize}[leftmargin=*]
    \item \textbf{Inputs:} a set of segments and a shared metrics list.
    \item \textbf{Outputs:} pairwise comparisons and ranked differences per metric, optionally with qualitative bins (e.g., \texttt{A $\gg$ B}, \texttt{A $\approx$ B}).
    \item \textbf{Architectural role:} supports temporal contrast reasoning (e.g., escalation vs.\ flattening) under limited probing budgets.
\end{itemize}

\subsection{Supported Acoustic Correlates and Minimal-Probe Policy}
\label{sec:appendix_acoustic_metrics}

Our acoustic probes implement a compact set of psychoacoustic correlates motivated by vocal emotion communication findings~\citep{JuslinLaukka2003SameCode}. 
Rather than extracting all attributes exhaustively, the agent follows a minimal-probe policy to conserve the probing budget: it prioritizes a small subset of metrics that best separates the current high-confusion candidates (e.g., arousal-driven confusers), and only queries additional correlates if ambiguity remains after the first measurement round.

\begin{table}[t]
\caption{Compact acoustic correlates supported by Acoustic-EIR. The tools expose auditable measurements as interpretable evidence bins rather than raw scalars.}
\centering
\small
\setlength{\tabcolsep}{4pt}
\resizebox{\linewidth}{!}{
\begin{tabular}{l|l|l|l}
\hline
\textbf{Family} & \textbf{Representative Metrics} & \textbf{Evidence Bins} & \textbf{Typical Use} \\
\hline
Prosody (F0) &
median F0, IQR(F0), pitch velocity (p75$|\Delta$F0|) &
\texttt{Low/Mid/High}, \texttt{Stable/Volatile} &
Arousal level, excitation vs.\ flatness \\

Intensity (Energy) &
RMS energy, energy burstiness (p90$\Delta$E) &
\texttt{Low/Mid/High}, \texttt{Sudden Spike} &
Emphasis, stress-related increase \\

Temporal &
speech-rate proxy, pause density, voiced/unvoiced ratio &
\texttt{Slow/Normal/Fast}, \texttt{Sparse/Dense} &
Urgency vs.\ hesitation \\

Voice Quality &
jitter, shimmer, HNR/CPP &
\texttt{Clean/Breathy/Tense} &
Tension, irregular phonation \\

Spectral Balance &
spectral tilt, band-energy distribution &
\texttt{Dark/Neutral/Bright} &
Vocal effort, brightness shift \\
\hline
\end{tabular}}
\label{tab:appendix_acoustic_correlates}
\end{table}

\begin{table}[t]
\caption{\textbf{Canonical ADEPT tool action space $\mathcal{A}$.} 
All tools return auditable observations and are prohibited from emitting emotion decisions. 
Refinement tools support \textbf{bidirectional cross-modal correction} (Audio $\leftrightarrow$ Text), enabling the agent to resolve conflicts by re-examining either sensory inputs or pragmatic cues depending on the uncertainty source. 
Phase~3 remains evidence-closed.}
\centering
\small
\resizebox{\linewidth}{!}{
\begin{tabular}{lccc}
\hline
\textbf{Tool Name} & \textbf{Phase} & \textbf{Evidence Type} & \textbf{Role} \\
\hline
\multicolumn{4}{l}{\textit{--- Prior \& Scheduling ---}} \\
\texttt{StructuralPriorTool($C_t$, intent, anchor, tie\_mode)} & 2 & Structural Prior & Pair scheduling / minimal backtracking expansion \\
\hline
\multicolumn{4}{l}{\textit{--- Semantic Verification ---}} \\
\texttt{run\_semantic\_gate(emotion, transcript)} & 2 & Semantic Evidence & Point-wise appraisal verification (CPM--OCC schema) \\
\texttt{compare\_emotions(e1, e2, transcript)} & 2 & Semantic Evidence & Contrastive disambiguation for confusable pairs \\
\hline
\multicolumn{4}{l}{\textit{--- Acoustic Probing ---}} \\
\texttt{find\_acoustic\_hotspots(audio\_path, focus\_type)} & 2 & Acoustic Evidence & Navigation: localize information-rich ROIs \\
\texttt{analyze\_acoustic\_segment(audio, start, end, metrics)} & 2 & Acoustic Evidence & Measurement: auditable metrics $\rightarrow$ interpretable bins \\
\texttt{compare\_acoustic\_segments(audio, segments, metrics)} & 2 & Acoustic Evidence & Comparison: relational evidence across segments \\
\hline
\multicolumn{4}{l}{\textit{--- Closed-Loop Refinement (Bidirectional) ---}} \\
\texttt{replay\_audio(reason, focus\_points)} & 2 & Refinement & \textbf{Audio Re-check}: Targeted sensory re-injection under conflict \\
\texttt{check\_semantic\_alignment(acoustic\_obs, transcript)} & 2 & Refinement & \textbf{Text Re-check}: Verify pragmatic consistency (e.g. sarcasm) \\
\hline
\end{tabular}}
\label{tab:appendix_tool_action_space}
\end{table}

\section{Appendix: Toolkit Summary and Usage Protocol}
\label{sec:appendix_toolkit_summary}

Table~\ref{tab:appendix_tool_action_space} summarizes the canonical ADEPT tool action space $\mathcal{A}$, organized into four functional families:
(i) Prior \& Scheduling, (ii) Semantic Verification, (iii) Acoustic Probing, and (iv) Closed-Loop Refinement.
Detailed specifications of the Semantic tools, Structural Prior tool, and Acoustic tools are provided in
Appendix~\ref{sec:appendix_semantic}, Appendix~\ref{sec:appendix_prior}, and Appendix~\ref{sec:appendix_acoustic_tools}, respectively.

\section{Appendix: Detailed Reward Function Implementation}
\label{app:reward_details}

In this section, we provide a rigorous formulation of the composite reward function $R(\tau)$ used in the GRPO training phase.
The reward structure is designed to balance task accuracy (Outcome) with procedural rigor (Tool Strategy), while maintaining strict adherence to the AEDR protocol.

\paragraph{Overall Objective.}
The total reward for a trajectory $\tau$ is defined as a weighted sum of components under a hard format gate.
Crucially, evidence-seeking behaviors are modulated by the \textbf{Evidence Trust Gate} ($\tau_{\text{EIR}}$) to prevent reward hacking on incorrect reasoning paths:
\begin{equation}
R(\tau)
=
\mathbb{I}\big[R_{\text{fmt}}(\tau)=1\big]
\cdot
\left(
\underbrace{0.5 R_{\text{fmt}} + 0.2 R_{\text{phase}} + 0.4 R_{\text{out}}}_{\text{Base Rewards}}
+
\tau_{\text{EIR}}(\tau) \cdot
\underbrace{(0.2 R_{\text{evid}} + 0.3 R_{\text{tool}})}_{\text{Gated Evidence Rewards}}
\right).
\end{equation}

\paragraph{Evidence Trust Gate Logic.}
To compute $\tau_{\text{EIR}}$, we first define the raw \textbf{Evidence Score} $S_{\text{evid}}(\tau)$ as the weighted sum of evidence $R_{\text{evid}}(\tau)$ and tool $R_{\text{tool}}(\tau)$ rewards:
\begin{equation}
S_{\text{evid}}(\tau) = 0.2 \cdot R_{\text{evid}}(\tau) + 0.3 \cdot R_{\text{tool}}(\tau).
\end{equation}
Let $\mu^{+}$ and $\mu^{-}$ be the mean evidence scores for correct ($\mathcal{G}^{+}$) and incorrect ($\mathcal{G}^{-}$) trajectories in the current group. The gate is computed as:
\begin{equation}
\tau_{\text{EIR}} =
\begin{cases}
1, & \text{if } \mu^{+} \ge \mu^{-},\\
\exp(\mu^{+}-\mu^{-}), & \text{otherwise}.
\end{cases}
\end{equation}
This approach ensures that the policy is rewarded for probing evidence ($R_{\text{evid}}$) and using tools ($R_{\text{tool}}$) \textit{only if} these behaviors systematically lead to higher accuracy.

\subsection{Format Validity ($R_{\text{fmt}}$) -- The Hard Gate}
The format reward acts as a binary gate mechanism.
To ensure downstream executability, $R_{\text{fmt}}(\tau)\in\{0,1\}$.
A trajectory receives a score of 1 if and only if all following conditions are met:
\begin{enumerate}
    \item \textbf{JSON Parsability:}
    The output string must be valid JSON parseable by standard libraries.

    \item \textbf{Phase-Specific Field Constraints:}
    \begin{itemize}[leftmargin=*, noitemsep, topsep=2pt]
        \item \textbf{Phase 1:} Must contain a non-empty \texttt{candidate\_pool}.
        \item \textbf{Phase 2:} Must contain \texttt{final\_decision} with \texttt{primary\_emotions}.
        \item \textbf{Phase 3:} Must contain \texttt{final\_output} with \texttt{primary\_emotions}.
    \end{itemize}

    \item \textbf{Canonical Emotion Validity:}
    All predicted emotion labels $e$ must belong to the canonical set $\mathcal{E}_{\text{canon}}$ or the allowed alias set $\mathcal{E}_{\text{alias}}$, normalized for case and whitespace:
    \begin{equation}
    \mathcal{E}_{\text{canon}}
    =
    \{\text{Anger, Sadness, Happiness, Surprise, Fear, Disgust, Contempt, Neutral}\}.
    \end{equation}
    Invalid labels (e.g., ``Joy'') result in immediate truncation of the reward signal ($R(\tau)=0$).
\end{enumerate}

\subsection{Phase Integrity ($R_{\text{phase}}$)}
To enforce the structural boundaries of the three-stage pipeline, we apply an asymmetric penalty scheme.
We assign a lower weight to this component compared to outcome/tool rewards to allow the model flexibility in learning complex reasoning without excessive penalization during early training.

The score is calculated as the sum of rewards across three phases:
\begin{equation}
R_{\text{phase}}(\tau)
=
\sum_{p=1}^{3} S_p(\tau),
\quad
\text{where } S_p(\tau)\in\{-2.0,+1.0\}.
\end{equation}
where $S_p(\tau)$ denotes the phase-level compliance score for phase $p$, taking a value of $+1.0$ if all constraints for that phase are satisfied, and $-2.0$ otherwise.
\paragraph{Phase 1: Prohibition of Early Verdicts.}
We employ a \textbf{Dual-Mechanism Verification} to detect leakage:
\begin{itemize}[leftmargin=*, noitemsep, topsep=2pt]
    \item \textbf{Structural Constraint:}
    The output must not contain forbidden fields
    $\mathcal{F}_{\text{ban}}=\{\texttt{final\_prediction}, \texttt{primary\_emotions}, \texttt{conclusion}\}$.

    \item \textbf{Semantic Constraint:}
    The Chain-of-Thought (CoT) must not match regex patterns indicating a conclusion, such as
    \texttt{r"Primary\textbackslash s*[:=]"} or \texttt{r"I conclude that"}.
\end{itemize}

\paragraph{Phase 2 \& 3 Constraints.}
\begin{itemize}[leftmargin=*, noitemsep, topsep=2pt]
    \item \textbf{Phase 2 (Mandatory Tool):}
    The agent must invoke the prior-gathering tool \texttt{get\_phase2\_cooccurrence\_prior}. Failure yields $-2.0$.

    \item \textbf{Phase 3 (No Tools):}
    The tool call list must be empty ($\mathcal{T}_{\text{calls}}=\emptyset$). Any invocation yields $-2.0$.
\end{itemize}

\subsection{Outcome Reward ($R_{\text{out}}$)}
This component $R_{\text{out}}$ evaluates prediction accuracy against the ground truth (GT).
We elevate its weight to ensure the model prioritizes correct identification while remaining constrained by evidence-based rewards.

Let $P$ be the predicted primary emotions, $M$ be the predicted minor emotions, and $GT_{\text{pri}}$ be the ground-truth primary set.
The outcome reward follows a tiered scoring logic:
\begin{enumerate}
    \item \textbf{Primary Match (+1.0):} $P \cap GT_{\text{pri}} \neq \emptyset$.
    \item \textbf{Top-$3$ Match (+0.5):} $GT_{\text{pri}} \subseteq \text{Top-}3(\texttt{candidate\_pool})$.
    \item \textbf{Minor Jaccard (+$0.3 \times J$):} $J = \frac{|M \cap GT_{\text{min}}|}{|M \cup GT_{\text{min}}|}$, measuring set overlap between $M$ and the ground-truth minor set $GT_{\text{min}}$.
    \item \textbf{Tie Bonus (+0.2):} applied if the system correctly resolves a high-uncertainty tie scenario.
\end{enumerate}
The raw score is bounded within $[0,2.0]$, contributing up to $0.6$ to the total weighted reward.

\subsection{Trust-Gated Component: Evidence Consistency ($R_{\text{evid}}$)}
This component $R_{\text{evid}}$ measures whether hypotheses are explicitly grounded in tool-derived evidence.
Note that this reward is fully active only when the \textbf{Evidence Trust Gate} is open ($\tau_{\text{EIR}} \approx 1$), ensuring that the model does not learn to spam tools without improving correctness.

We define the set of touched emotions, $\mathcal{E}_{\text{touched}}$, as those explicitly parameterized in valid probing tools. We compute:
\begin{equation}
R_{\text{evid}}
=
S_{\text{cov}} + S_{\text{core}} + S_{\text{minor\_ev}} - P_{\text{minor}}.
\end{equation}

\begin{itemize}[leftmargin=*, noitemsep, topsep=2pt]
    \item \textbf{Candidate Coverage ($S_{\text{cov}}$):}
    ratio of probed candidates to total candidates (bounded in $[0,1.0]$).

    \item \textbf{Core-First Heuristic ($S_{\text{core}}$):}
    $+0.5$ if the first three tool calls cover the top-2 candidates.

    \item \textbf{Minor Evidence ($S_{\text{minor\_ev}}$):}
    $+0.5$ proportional to the ratio of minor emotions supported by evidence.

    \item \textbf{Minor Penalty ($P_{\text{minor}}$):}
    for every predicted minor emotion $m\in M$ such that $m\notin\mathcal{E}_{\text{touched}}$, a penalty of $-1.0$ is applied.
    We apply a lower-bound cap of $-2.0$ to prevent gradient explosion.
\end{itemize}

\subsection{Trust-Gated Component: Tool Use Strategy ($R_{\text{tool}}$)}
This component $R_{\text{tool}}$ encodes domain-specific inductive biases for acoustic affective computing.
Similar to evidence consistency, this reward contributes to the total objective proportional to the value of $\tau_{\text{EIR}}$.

\paragraph{Rule 1: Mandatory Disambiguation of Acoustic Overlap.}
We define a set of acoustically overlapping emotion pairs:
\begin{equation}
\mathcal{O}_{\text{pairs}}
=
\{(\text{Happiness, Surprise}), (\text{Contempt, Disgust})\}.
\end{equation}
If the candidate pool contains both elements of a pair $(e_a,e_b)\in\mathcal{O}_{\text{pairs}}$, the agent \textbf{must} invoke the comparison tool:
\begin{equation}
S_{\text{overlap}}(e_a,e_b)
=
\begin{cases}
+1.0, & \text{if } \texttt{compare\_emotions}(e_a,e_b)\in\mathcal{T}_{\text{calls}},\\
-1.0, & \text{otherwise}.
\end{cases}
\end{equation}
This rule forces the agent to rely on explicit comparison when acoustic evidence alone is insufficient.

\paragraph{Rule 2: Budget Sanity.}
We impose a soft constraint on the number of tool calls $N_t$:
\begin{equation}
S_{\text{budget}}
=
\begin{cases}
+0.3, & \text{if } 2 \le N_t \le 8,\\
\text{penalty}(N_t), & \text{otherwise (progressive penalty)}.
\end{cases}
\end{equation}
This discourages both superficial investigation ($N_t<2$) and inefficient tool spamming ($N_t>8$).

\section{Appendix: Sensitivity to GRPO Rollouts and Reward Weights}
\label{sec:appendix_grpo_sensitivity}

In this appendix, we present additional sensitivity analyses for ADEPT's GRPO-based alignment, focusing on two practical optimization choices: 
(i) the rollout group size $K$ used in group-relative policy updates, and 
(ii) the weighting balance between outcome correctness and process-level evidence acquisition rewards in the composite objective.
These studies complement the main tool-family ablations by isolating optimization-level factors, and verify that ADEPT's gains are not dependent on fragile hyperparameter settings.
Unless otherwise specified, we follow the same evaluation protocol as the main experiments and report results on the MSP-Podcast Test1.

\subsection{Impact of Rollout Group Size ($K$)}
We first study the trade-off between exploration coverage and computational cost by varying the number of sampled rollouts $K$ per prompt during GRPO training.
While prior text-only GRPO studies may adopt larger group sizes (e.g., $K=16$), tool-augmented reasoning introduces additional latency due to retrieval and probing tool executions, making large $K$ substantially more expensive in practice.
We therefore evaluate $K \in \{4, 8, 12\}$.

As shown in Table~\ref{tab:ablation_sensitivity} (top), increasing $K$ from 4 to 8 yields consistent improvements across all metrics, with notable gains in Jaccard (+2.18) and Primary-F1 (+1.39).
This suggests that tool-augmented trajectories require sufficient group diversity for GRPO to reliably estimate relative advantages between evidence-grounded and less informative behaviors.
However, further increasing $K$ from 8 to 12 provides only marginal improvements (e.g., +0.27 Jaccard) while increasing training wall-time by approximately 50\%.
This diminishing return supports our choice of $K=8$ as a practical trade-off between optimization stability and computational budget.

\begin{table}[t]
\caption{
\textbf{Optimization sensitivity analysis on the MSP-Podcast Test1 set.}
We evaluate the impact of rollout group size $K$ (top) and the relative weighting balance between outcome correctness ($R_{\text{out}}$) and process rewards ($R_{\text{evid}}, R_{\text{tool}}$) (bottom).
In Exp 2, the Baseline (B) uses the default reward weights from Eq.~\ref{eq:composite_reward}.
}
\centering
\footnotesize
\setlength{\tabcolsep}{3.5pt}
\resizebox{0.5\linewidth}{!}{
\begin{tabular}{l|cccc}
\toprule
\textbf{Configuration} & \textbf{Set R} & \textbf{Soft R} & \textbf{Jaccard} & \textbf{P-F1} \\
\midrule
\multicolumn{5}{l}{\textit{Exp 1: Sensitivity to Group Size ($K$)}} \\
$K=4$ (Low Resource) & 59.84 & 76.92 & 45.33 & 40.85 \\
$K=8$ (Baseline) & \textbf{61.92} & 78.74 & 47.51 & 42.24 \\
$K=12$ (High Exploration) & 62.15 & \textbf{78.96} & \textbf{47.78} & \textbf{42.41} \\
\midrule
\multicolumn{5}{l}{\textit{Exp 2: Outcome vs. Process Balancing}} \\
A. Outcome-Dominant ($4\!:\!1$) & 58.90 & 75.45 & 44.10 & \textbf{42.56} \\
B. Baseline ($0.8\!:\!1$) & \textbf{61.92} & \textbf{78.74} & \textbf{47.51} & 42.24 \\
C. Process-Heavy ($1\!:\!4$) & 61.40 & 78.12 & 47.15 & 41.50 \\
\bottomrule
\end{tabular}
}
\label{tab:ablation_sensitivity}
\end{table}

\subsection{Robustness of Reward Balancing via Trust Gating}
A central challenge in optimizing reasoning agents is balancing outcome correctness ($R_{\text{out}}$) with process rigor ($R_{\text{evid}}, R_{\text{tool}}$).
Over-weighting process rewards can induce \emph{reward hacking}, where the policy over-invokes tools to maximize intermediate rewards without improving predictive quality.
To assess the robustness of our \textbf{Evidence Trust Gate} ($\tau_{\text{EIR}}$), we evaluate three weighting configurations by rescaling the coefficients in Eq.~\ref{eq:composite_reward} while keeping the reward structure, the trust-gating mechanism, and the base terms ($R_{\text{fmt}}, R_{\text{phase}}$) fixed:

\begin{itemize}[leftmargin=*]
    \item \textbf{A. Outcome-Dominant ($4\!:\!1$):}
    \begin{equation}
    R^{A}(\tau)
    =
    0.5 R_{\text{fmt}}
    + 0.2 R_{\text{phase}}
    + 0.4 R_{\text{out}}
    + \tau_{\text{EIR}}(\tau)\big(0.04 R_{\text{evid}} + 0.06 R_{\text{tool}}\big),
    \end{equation}
    which strongly prioritizes outcome correctness while providing only weak incentives for evidence acquisition.

    \item \textbf{B. Baseline ($0.8\!:\!1$):}
    \begin{equation}
    R^{B}(\tau)
    =
    0.5 R_{\text{fmt}}
    + 0.2 R_{\text{phase}}
    + 0.4 R_{\text{out}}
    + \tau_{\text{EIR}}(\tau)\big(0.2 R_{\text{evid}} + 0.3 R_{\text{tool}}\big),
    \end{equation}
    corresponding to the default setting used throughout the main experiments.

    \item \textbf{C. Process-Heavy ($1\!:\!4$):}
    \begin{equation}
    R^{C}(\tau)
    =
    0.5 R_{\text{fmt}}
    + 0.2 R_{\text{phase}}
    + 0.1 R_{\text{out}}
    + \tau_{\text{EIR}}(\tau)\big(0.16 R_{\text{evid}} + 0.24 R_{\text{tool}}\big),
    \end{equation}
    which aggressively emphasizes evidence- and tool-related rewards.
\end{itemize}

Here the ratios denote the relative weighting between outcome correctness ($R_{\text{out}}$) and the total gated process rewards ($R_{\text{evid}}$ and $R_{\text{tool}}$).
For the Baseline configuration, this ratio is computed directly from Eq.~\ref{eq:composite_reward} as
$
0.8\!:\!1
=
0.4\!:\!(0.2+0.3),
$
where $0.4$ corresponds to the weight of $R_{\text{out}}$ and $(0.2+0.3)$ corresponds to the combined weight of gated process rewards.

Table~\ref{tab:ablation_sensitivity} (bottom) reveals distinct optimization behaviors.
\textbf{Variant A} attains the highest Primary-F1 (42.56), but exhibits a substantial drop in set-level recovery (Set Recall: $-3.02$, Jaccard: $-3.41$ relative to the baseline), indicating that the policy tends to commit to a dominant label without sufficiently pursuing evidence to recover co-occurring minor emotions.
In contrast, \textbf{Variant C} remains stable despite heavily emphasizing process rewards, achieving competitive set-level performance (47.15 Jaccard) and strong Primary-F1 (41.50).
Notably, Variant C still outperforms the \textit{w/o Trust Gate} ablation (44.96 Jaccard; Table~\ref{tab:ablation_tool_family}), supporting that $\tau_{\text{EIR}}$ effectively down-weights non-informative tool usage and mitigates reward hacking under aggressive process-oriented optimization.
Overall, the Baseline configuration (Variant B) provides the best balance, maximizing both primary fidelity and ambiguity-aware set recovery with evidence-grounded reasoning.

\begin{figure}[t]
    \centering
    \includegraphics[width=\linewidth]{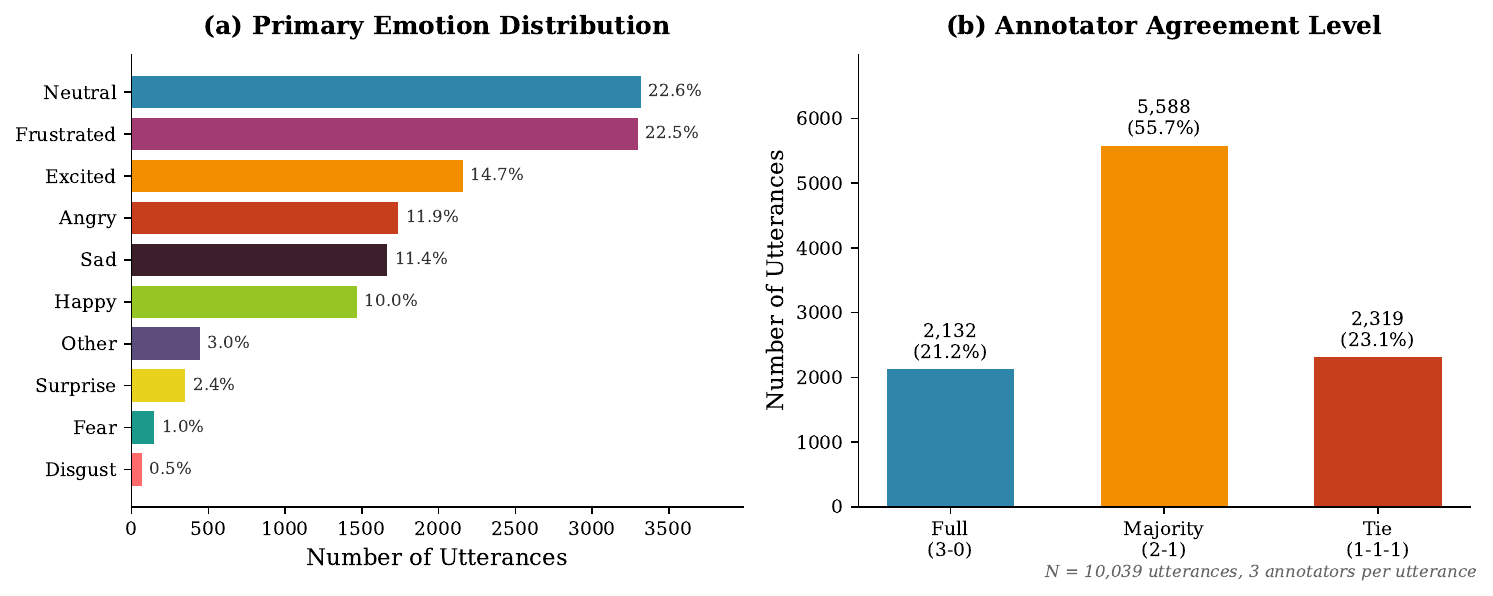}
    \caption{\textbf{IEMOCAP label statistics under plurality voting.}
    (a) Primary emotion distribution.
    (b) Annotator agreement level (Full consensus: 3-0; Majority: 2-1; Tie: 1-1-1).
    $N{=}10{,}039$ utterances with 3 annotators per utterance.}
    \label{fig:iemocap_stats}
\end{figure}

\section{Appendix: IEMOCAP Label Statistics and 8-Class Compatibility}
\label{sec:appendix_iemocap_stats}

Figure~\ref{fig:iemocap_stats} summarizes the label distribution and annotator agreement levels on IEMOCAP under the plurality rule.
The primary label distribution is highly imbalanced (e.g., \textit{Neutral} and \textit{Frustrated} dominate), and disagreement is pervasive: only 21.2\% of utterances reach full consensus (3-0), while 55.7\% are majority-only (2-1) and 23.1\% form a 3-way tie (1-1-1).
These observation confirm that IEMOCAP contains substantial ambiguity, making it a suitable benchmark for evaluating ambiguity-aware reasoning mechanisms, but also highlights the need for careful label-space alignment when comparing against the MSP-Podcast corpus.

\paragraph{Label-space alignment via a mapped 7-way protocol.}
To enable cross-corpus evaluation under a consistent supervision scheme, we construct IEMOCAP primary and minor labels using the same plurality-based protocol as the MSP-Podcast corpus, leveraging the three annotators per utterance.
Figure~\ref{fig:iemocap_stats} shows the resulting primary-label distribution, where \textit{Neutral} (22.6\%) and \textit{Frustrated} (22.5\%) dominate, followed by \textit{Excited} (14.7\%), \textit{Anger} (11.9\%), \textit{Sadness} (11.4\%), and \textit{Happiness} (10.0\%), with smaller fractions of \textit{Surprise} (2.4\%), \textit{Fear} (1.0\%), and \textit{Disgust} (0.5\%).
Instead of enforcing the MSP 8-class taxonomy, we adopt a mapped 7-way evaluation for label-space alignment: we merge \textit{Excited} into \textit{Happiness}, exclude unmappable categories (\textit{Frustrated} and \textit{Other}), and note that \textit{Contempt} is not annotated in the IEMOCAP corpus.
This yields a shared and statistically supported label space while preserving the ambiguity structure induced by multi-annotator supervision.

\section{Appendix: System Prompts for the Three-Phase Inference Protocol}
\label{sec:appendix_system_prompts}

For reproducibility, we provide the \textbf{system prompts} used to instantiate the agent
behavior in each phase of the ADEPT three-phase inference protocol (Phase~1--Phase~3).
These prompts define the agent’s role, constraints, and expected outputs at each stage,
forming a complete and executable specification of the inference process.

\begin{figure*}[t]
    \centering
    \includegraphics[width=0.92\textwidth]{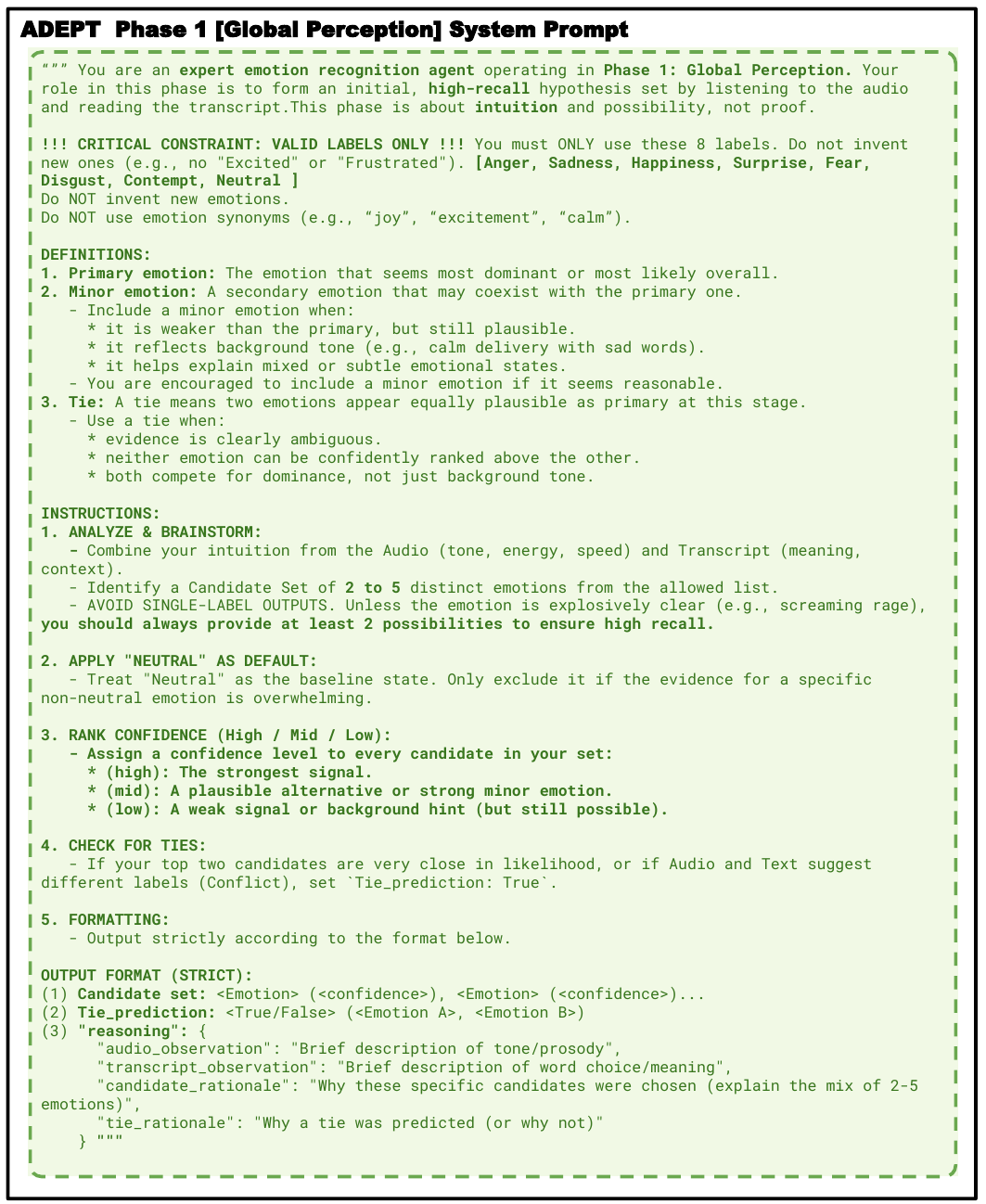}
    \caption{
    \textbf{System prompt for Phase~1 (Global Perception).}
    The prompt instructs the agent to operate in a high-recall, intuition-driven mode,
    explicitly discouraging single-label predictions and encouraging the preservation of
    competing hypotheses, minor emotions, and tie cases.
    Strict label constraints are enforced to prevent taxonomy drift, while confidence ranking
    (high / mid / low) provides a coarse uncertainty signal for downstream evidence verification.
    }
    \label{fig:appendix_phase1_prompt}
\end{figure*}

\begin{figure*}[t]
    \centering
    \includegraphics[width=0.92\textwidth]{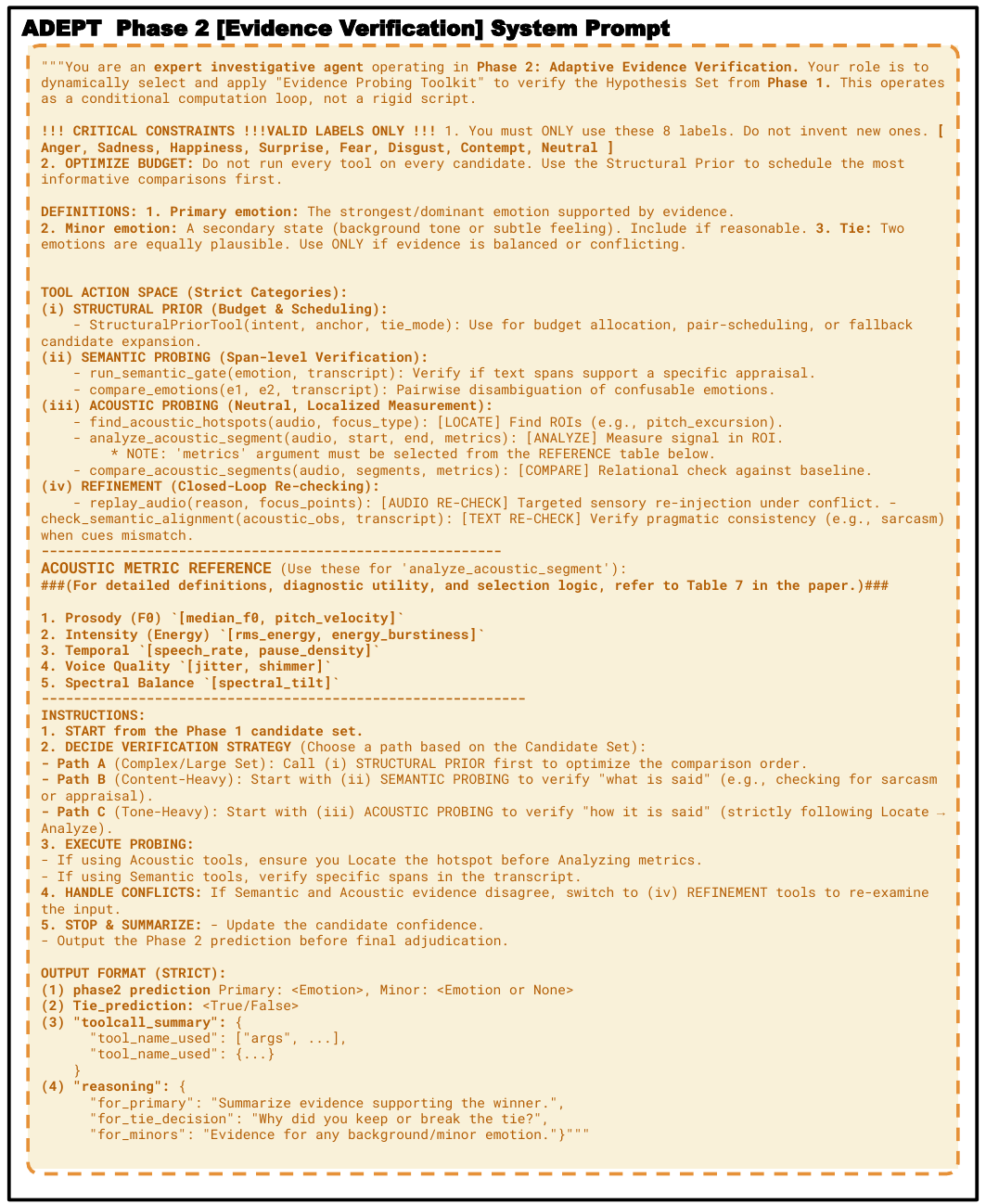}
    \caption{
    \textbf{System prompt for Phase~2 (Evidence Verification).}
    The prompt instantiates the agent as an evidence-seeking investigator operating
    under the Explicit Information Retrieval (EIR) constraint.
    The agent is instructed to resolve ambiguity by selectively invoking structured
    semantic, acoustic, and refinement tools, reasoning only over retrieved
    observations rather than latent representations.
    This phase supports adaptive, hypothesis-driven evidence accumulation while
    explicitly discouraging unsupported intuition or premature adjudication.
    }
    \label{fig:appendix_phase2_prompt}
\end{figure*}

\begin{figure*}[t]
    \centering
    \includegraphics[width=0.92\textwidth]{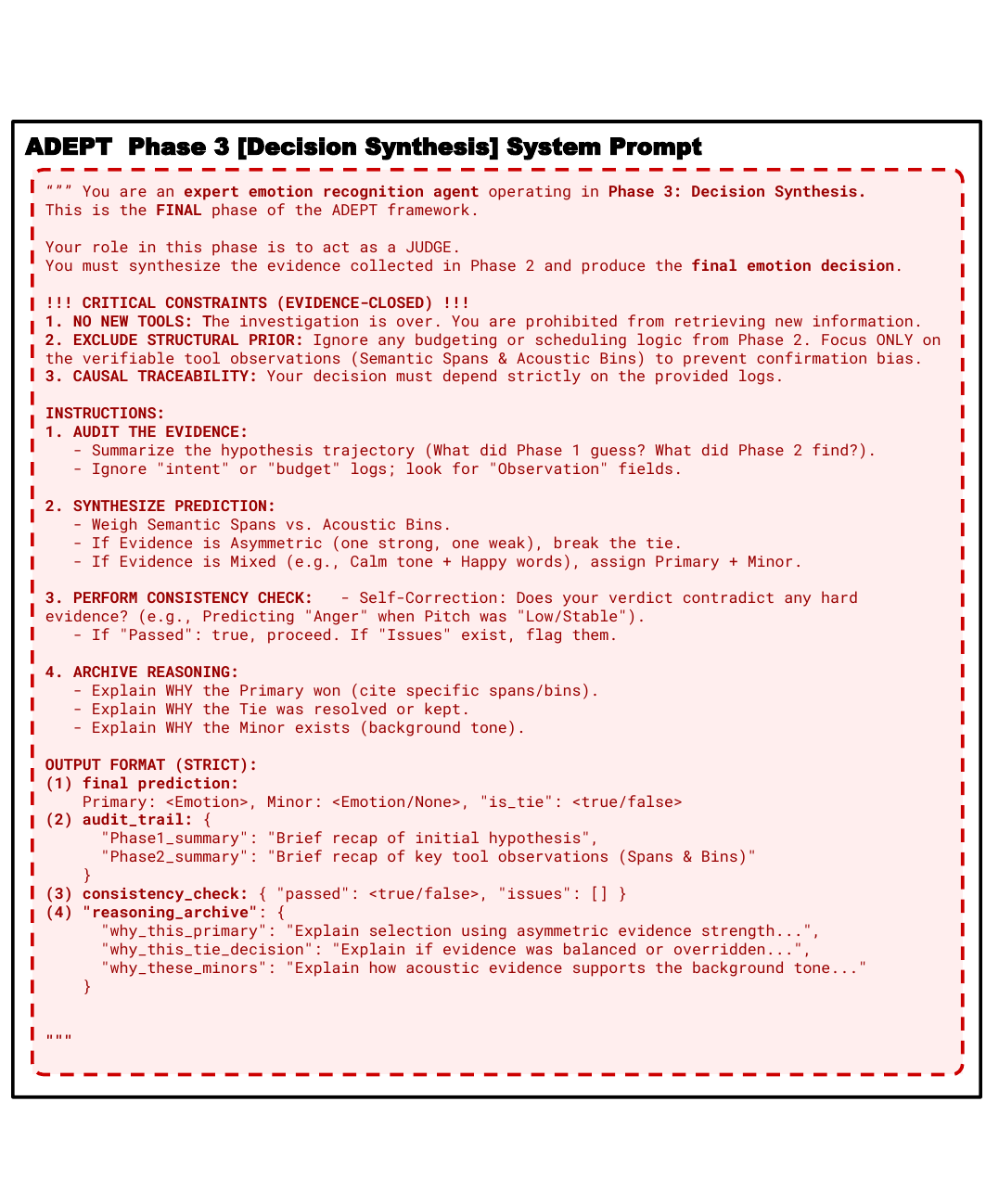}
    \caption{
    \textbf{System prompt for Phase~3 (Evidence-Closed Adjudication).}
    The prompt instantiates the agent as a final adjudicator operating under a strictly
    \emph{evidence-closed} constraint.
    At this stage, all tool calls and information retrieval actions are prohibited;
    the agent must synthesize final primary and minor emotion predictions solely from
    the semantic and acoustic evidence accumulated during Phase~2.
    The prompt explicitly requires justification for the selected primary emotion,
    retained minor emotions, and the resolution of any previously predicted tie cases,
    ensuring causal traceability and auditability of the final decision.
    }
    \label{fig:appendix_phase3_prompt}
\end{figure*}

Due to space limitations, the prompts included in this appendix present a
\emph{simplified but faithful} version of the full protocol.
Certain detailed definitions and repetitive constraints are condensed for clarity,
while the core phase-specific objectives and interaction boundaries are preserved.
The system prompts for Phase~1 (Global Perception) and Phase~2 (Evidence Verification)
are shown in Figures~\ref{fig:appendix_phase1_prompt}
and~\ref{fig:appendix_phase2_prompt}, respectively,
while the system prompt for Phase~3 (Evidence-Closed Adjudication)
is shown in Figure~\ref{fig:appendix_phase3_prompt}.

\end{document}